\documentclass[journal,10pt]{IEEEtran}
\ifCLASSINFOpdf
\else
\fi
\usepackage[]{graphicx}
\usepackage[sf,SF]{subfigure}
\usepackage{psfrag}
\usepackage[pdf]{pstricks}
\usepackage{epsfig}
\usepackage{amsthm}
\usepackage{amsmath}
\usepackage{amssymb}
\usepackage{float}
\usepackage{dsfont}
\usepackage{multirow}
\usepackage{bm}
\usepackage{array}
\usepackage{fancyhdr}
\usepackage{booktabs} % make nicer tables
\usepackage{bbm}
\usepackage{cite}
\usepackage{rotating}
\usepackage{hyperref}

\newcommand{\ra}[1]{\renewcommand{\arraystretch}{#1}}

\newlength{\noteWidth}
\long\def\notes#1{\ifinner
             {\tiny #1}
             \else
              \marginpar{\parbox[t]{\noteWidth}{\raggedright\tiny #1}}
               \fi}

\newcommand{\norm}[1]{\left\Vert#1\right\Vert}

\def\fparam2{{f_{\underline{\alpha_1,\ldots,\alpha_\pop}}}}

\def\F_theta{F_\theta}

\DeclareMathOperator*{\argmax}{arg\,max}
\DeclareMathOperator*{\argmin}{arg\,min}

\def\prdis{\pi}
\def\prdisvec{\bm{\pi}}

\def\staten{K}
\def\timen{T}
\def\pop{N}

\def\llikeli{\mathcal{L}}

\def\prsim{\mathcal{P}}

\def\bigO{\mathcal{O}}	%big o notation

\def\z{j}
\def\j{m}

\def\zstate{\mathcal{S}}  %   \def\zstate{{\sf Z}} $\SSet$
\def\clI{{\cal I}}

\def\clH{{\cal H}}
\def\FRAC#1#2#3{\genfrac{}{}{}{#1}{#2}{#3}}

\def\half{{\mathchoice{\FRAC{1}{1}{2}}%
{\FRAC{1}{1}{2}}%
{\FRAC{3}{1}{2}}%
{\FRAC{3}{1}{2}}}}

\def\what{\widehat}

\def\SetSa{\psi_{1}}
\def\SetSb{\psi_{2}}

\def\SetSaTilde{\widetilde{\psi}_{1}}

\def\USetSa{U_1}
\def\USetSb{U_2}

\def\permU{\bm{\sigma}}

\usepackage{ifthen}

\newcounter{example}

\newcommand{\appropto}{\mathrel{\vcenter{
  \offinterlineskip\halign{\hfil$##$\cr
    \propto\cr\noalign{\kern2pt}\sim\cr\noalign{\kern-2pt}}}}}

\begin{document}
%\title{Histograms as Fingerprints: User Identification by Matching Statistics}
\title{Where You Are Is Who You Are: User Identification by Matching Statistics}

\author{Farid~M.~Naini, Jayakrishnan~Unnikrishnan,~\IEEEmembership{Member,~IEEE}, Patrick~Thiran,~\IEEEmembership{Fellow, IEEE,}        
and~Martin~Vetterli,~\IEEEmembership{Fellow,~IEEE}% <-this % stops a space
\IEEEcompsocitemizethanks{\IEEEcompsocthanksitem Copyright (c) 2013 IEEE. Personal use of this material is permitted. However, permission to use this material for any other purposes must be obtained from the IEEE by sending a request to pubs-permissions@ieee.org.
\IEEEcompsocthanksitem  F. M. Naini,  J. Unnikrishnan, P. Thiran, and M. Vetterli are with EPFL, Lausanne, Switzerland.
E-mail: farid.naini@alumni.epfl.ch, dr.j.unnikrishnan@ieee.org,  patrick.thiran@epfl.ch, and  martin.vetterli@epfl.ch.%\protect\\
%E-mail: see http://www.michaelshell.org/contact.html
%\thanks{Manuscript received April 19, 2005; revised September 17, 2014.}
}
}

\maketitle

%\begin{abstract}
%
%Must mention a new distance metric to compare histograms!
%Our approach is domain-independent!
%
%%We analyze the vulnerability of  anonymized mobility statistics stored in the form of histograms.
%%We consider an attacker who has access to an anonymized set of histograms of a set of  users' mobility traces and to an \emph{independent} set of non-anonymized histograms of traces belonging to the same users.
%%An example of independent set of histograms  is the case where the anonymized histograms and the non-anonymized histograms (also called auxiliary histograms) are collected over two non-overlapping time periods.
%%We study the hypothesis-testing problem of identifying the correct matching between the anonymized histograms and the auxiliary histograms.
%%We consider the problem from the perspective of the attacker, and we consider the case where only the user identities are removed from the anonymized histograms as well as the case where both the user identities and the location identities are removed.
%%We investigate how much the additional removal of location identities complexifies the de-anonymization.
%%By applying our proposed algorithms to real mobility traces, we demonstrate that anonymized mobility statistics  contain a significant amount of information that could be used to uniquely identify users by an attacker having access to auxiliary information about the statistics.
%\end{abstract}

\begin{abstract}
Most users of online services have unique behavioral or usage patterns.
These behavioral patterns can be exploited to identify and track users by using only the observed patterns in the behavior.
We study the task of identifying users from statistics of their behavioral patterns.
Specifically, we focus on the setting in which we are given histograms of users' data collected during two different experiments.
%In the first dataset, we assume that the users' identities are anonymized or hidden and in the second dataset we assume that their identities are known.
We assume that, in the first dataset, the users' identities are anonymized or hidden and that,  in the second dataset, their identities are known.
We study the task of identifying the users by matching the histograms of their data  in the first dataset with the histograms from the second dataset.
In  recent works~\cite{unniMNainiAlert13,unn13} the optimal algorithm for this user identification task is introduced.
In this paper, we evaluate the effectiveness of this method on three different types of  datasets with up to $50,000$ users, and in multiple scenarios.
Using datasets such as call data records, web browsing histories, and GPS trajectories, we demonstrate that a large fraction of users can be easily identified given only histograms of their data; hence these histograms can act as users' \emph{fingerprints}.
We also verify  that  simultaneous identification of  users achieves better performance compared to  one-by-one user identification.
Furthermore, we show that  using the optimal method for identification indeed gives higher identification accuracy than heuristics-based approaches in  practical scenarios.
The accuracy obtained under this optimal method can thus be used to quantify the maximum level of user identification that is possible in such settings.
We show that the key factors affecting the accuracy of the optimal identification algorithm are the duration of the data collection, the number of users in the anonymized dataset, and the resolution of the dataset.
We also analyze the effectiveness of $k$-anonymization in resisting user identification attacks on these datasets.
%\footnote{Following the principle of \emph{reproducible research}, the code and the data used in this paper are made available online at \url{rr.epfl.ch}.}
\footnote{Following the principle of \emph{reproducible research}, the code for performing user matching and for generating the figures related to the publicly available datasets are made available for download at \url{rr.epfl.ch}.}
\end{abstract}


% \fi
\IEEEpeerreviewmaketitle

%\section{Introduction}

%%%%%%%%%%%%%%%%%%%%%%%%%%%%%%%%%%%%%%%%%%%%%%%%%%%%%%%%%%%%%%%%%%%%%%%%%%%%%%%%%%%
%%%%%%%%%%%%%%%%%%%%%%%%%%%%%%%%%%%%%%%%%%%%%%%%%%%%%%%%%%%%%%%%%%%%%%%%%%%%%%%%%%%
%%%%%%%%%%%%%%%%%%%%%%%%%%%%%%%%%%%%%%%%%%%%%%%%%%%%%%%%%%%%%%%%%%%%%%%%%%%%%%%%%%%
%%%%%%%%%%%%%%%%%%%%%%%%%%%%%%%%%%%%%%%%%%%%%%%%%%%%%%%%%%%%%%%%%%%%%%%%%%%%%%%%%%%
%%%%%%%%%%%%%%%%%%%%%%%%%%%%%%%%%%%%%%%%%%%%%%%%%%%%%%%%%%%%%%%%%%%%%%%%%%%%%%%%%%%

%%%%%%%%%%%%%%%%%%%%%%%%%%%%%%%%%%%%%%%%%%%%%%%%%%%%%%
%%%%%%%%%%%%%%%%%%%%%%%%%%%%%%%%%%%%%%%%%%%%%%%%%%%%%%
%%%%%%%%%%%%%%%%%%%%%%%%%%%%%%%%%%%%%%%%%%%%%%%%%%%%%%
\section{Introduction}\label{sec:intro}
A common task in data analysis is to identify users by exploiting statistics of their data.
%In many applications, we have access to some information, from two sources, about two sets of the same unidentified users, and the task is to \emph{match} pieces of this information that  correspond to the same underlying users from the first source to those  from the second.
%If the identities of the users in these two sets are known, then this is a trivial task.
In many applications, we have access to some information about a set of users from one source, and some other information about the set of users from another source, and the task is to \emph{match} pieces of information from the first source to pieces of information from the second source that correspond to the same underlying user.
If the identities of the users in the two sets are known, then this is a trivial task.
However, in many practical applications, the identities of the users are unknown either in the first set or in the second set or in both; therefore,  in such cases, the task becomes non-trivial.
For example, the two datasets might contain information about location statistics of users in a city measured over distinct time periods.
The first set can be obtained by tracking cell phone connections to cell-towers and the second set can be obtained from credit-card usage patterns.
In this case, the  task is to identify the correct matching from the users' phone numbers to their credit card numbers.
Another example of the matching problem is relevant to datasets collected from the same service during two different time periods.
For instance, the users on a website might choose to change their online user identities for privacy purposes; but given the statistics of the users' data measured prior to the change and after the change, it might still be possible to identify the users by matching the statistics across the two time periods.
Matching users between two datasets increases the net information available about the users, which in turn is useful for providing better targeted advertisements and recommendations for products and services~\cite{resvar97}.
%[RRRR: more references and also reference to mix zones]
%
%Another example might be shopping patterns of a set of users measured across two different time-periods.
%
%Other applications include matching of shopping patterns of users across
%
%
%
%If the identities of the users are not known on one of the two days, the
%
%
%
%A more challenging example is to match users
%
%For example, given browsing patterns of users measured on
%
%An example for such a task on online data is to identify profiles belonging to the same individuals in different social networks, e.g., Facebook and LinkedIn, or two different websites, e.g., Amazon and eBay.
%This task becomes non-trivial if multiple users share the same profile names.
%

The problem of matching users is also relevant in the context of privacy of an anonymized database.
In recent years, many datasets containing information about individuals have been released into the public domain in order to provide open access to statistics or to facilitate data-mining research.
Often these databases are \emph{anonymized} by suppressing identifiers that reveal the identities of the users, such as names or social security numbers.
Nevertheless, recent research has revealed that the privacy offered by such anonymized databases could be compromised, if an adversary correlates the revealed information with auxiliary information about the users from publicly available databases.
A famous example of such a de-anonymization attack was shown in~\cite{Narayanan:2008:RDL:1397759.1398064}, in which anonymous movie ratings released during the Netflix Prize contest were de-anonymized by using public user reviews from the Internet Movie Database (IMDB).
In such attacks, the adversary's task of de-anonymization  is essentially a matching task.
The objective is to identify users in the anonymized dataset by matching their data to the publicly available auxiliary information.

As the question of matching users is relevant in many applications, this problem has been studied by many authors in different fields, including database management~\cite{elmagarmid2007duplicate}, information retrieval~\cite{kalashnikov2008web}, natural language processing~\cite{bengtson2008understanding}, author identification~\cite{stolerman2014breaking,afroz2014doppelganger}, and privacy ~\cite{Narayanan:2008:RDL:1397759.1398064}.
Nevertheless, most solutions to the matching problem rely on heuristics that are relevant for specific applications, but not for other applications.
In this paper, we present a systematic study of the matching problem under a general setting.
The problem we study  differs from typical approaches in data analysis in that we focus on the setting in which the available information  about a user's data is in the form of histograms of the user's data.
The histograms capture the habits of the users.
In the case of mobility traces,  such histograms could be the average time spent by each user at different locations during a day, or during different time intervals.
In some applications, such as urban planning, the data collected naturally contains only the statistics of the data, as they are sufficient for such applications.
In  other applications, the data is intentionally stripped of timing information to enhance the privacy of the users; in which case, all that remains are histograms.
%For example, in the case of location data, the histograms of the users' data represent the fraction of time that the users spend in various locations.%, or in the case of browsing data, the fraction of visits to various websites.
We study the problem of matching histograms of users' data measured in two independent experiments
%.
%We introduce the idea of studying this problem
as a hypothesis testing problem.
This novel formulation has the advantage of making it  possible to rigorously define the accuracy of a matching scheme and to identify an algorithm that is provably more accurate than other schemes.

An example of a user identification task, is to consider a dataset comprised of unlabeled location histograms, given in Table~\ref{tab: Ch4illustrativeexp1}\subref{tab: Ch4illustrativeexp1a}, where the user identities are removed.
Now consider an adversary who has access to the labeled location histograms of the same users  in an independent experiment where the user identities are known (refer to Table~\ref{tab: Ch4illustrativeexp1}\subref{tab: Ch4illustrativeexp1b}).
This information could be obtained, for instance, by tracking the users during a
%different time-period.
%Furthermore, the anonymized histograms in Tables~\ref{tab: Ch4illustrativeexp1}\subref{tab: Ch4illustrativeexp1b} are collected over a
different time-period compared  to those in Table~\ref{tab: Ch4illustrativeexp1}\subref{tab: Ch4illustrativeexp1a}.
The histograms corresponding to  each user in the two tables are expected to be similar, as the habits of the user are expected to remain the same across the two datasets; but they might not be exactly identical due to the inherent randomness in the user's behavior.
The objective of the adversary is to match the user identities (i.e., the rows) across the two tables.
\begin{table}
\addtolength{\subfigcapskip}{5pt}
        \centering
	\ra{1.2}
	\setlength{\belowrulesep}{0pt}
	\setlength{\aboverulesep}{0pt}
	\subtable[Unlabeled histograms (Day $1$)]{
	\begin{footnotesize}
	\begin{tabular}{|c|c|c|c|}
			\cmidrule[1pt]{1-4}
			\!\!User\!\! & \multicolumn{3}{c|}{Location}  \tabularnewline
		\cmidrule{2-4}
	 & \!\!Dorm.\!\! & \!\!Rest.\!\! &  \!\!Lib.\!\!  \tabularnewline
		\midrule[0.5pt]
	 ? & $75\%$   & $15\%$  &  $10\%$  \tabularnewline
		\cmidrule{1-4}
		 ? & $31\%$  & $30\%$  &  $39\%$   \tabularnewline
		\cmidrule{1-4}
		 ? & $15\%$  & $15\%$  &  $70\%$    \tabularnewline
		\cmidrule{1-4}
		 ? & $15\%$  & $65\%$  &  $20\%$    \tabularnewline
		\bottomrule[1pt]
		\end{tabular}
	\end{footnotesize}
%			\end{tiny}
     \label{tab: Ch4illustrativeexp1a}
     }
    \hfill
    \subtable[Labeled histograms (Day $2$)]{
	\begin{footnotesize}
	\begin{tabular}{|c|c|c|c|}
			\cmidrule[1pt]{1-4}
			\!\!User\!\!  & \multicolumn{3}{c|}{Location}  \tabularnewline
		\cmidrule{2-4}
	 & \!\!Dorm.\!\! & \!\!Rest.\!\! &  \!\!Lib.\!\!  \tabularnewline
		\midrule[0.5pt]
	 John & $33\%$   & $33\%$  &  $34\%$  \tabularnewline
		\cmidrule{1-4}
		 Jill & $70\%$  & $20\%$  &  $10\%$   \tabularnewline
		\cmidrule{1-4}
		 Mary & $15\%$  & $60\%$  &  $25\%$    \tabularnewline
		\cmidrule{1-4}
		 Mike & $15\%$  & $20\%$  &  $65\%$    \tabularnewline
		\bottomrule[1pt]
		\end{tabular}	
\end{footnotesize}
    \label{tab: Ch4illustrativeexp1b}
    }
\caption{An illustrative example of user identification task  on  histograms. Location statistics in the form of histograms of some users are released (in (a)), where the user identities are removed. An adversary has access to some auxiliary histograms (in (b)) about the same users where the user identities are known.
The time period during which the histograms in (a) are collected does not overlap with the time period during which the histograms in (b) are collected. %Therefore,  for each user, the corresponding histograms in (a) and (b) may not be exactly the same.
    The objective of the adversary is to match the users (i.e., rows) across the two tables.}
    \label{tab: Ch4illustrativeexp1}
    \vspace{-0.5cm}
\end{table}

%The rest of this paper is organized as follows.
In the next section, we provide a detailed comparison of this problem  with existing literature on user identification and highlight the new contributions of this work.
We state the problem in mathematical form and propose our solution in Section~\ref{sec: Ch4problemstatement}.
%We present an extension of our solution to a more general scenario in Section~\ref{sec: uncbipartitematch1}.
%We also discuss the case where in addition to histograms, some further information about the users might be available.
We   experimentally evaluate our solution by using three different datasets in Section~\ref{sec: experiments}. %a Call Data Records dataset in Section[?] and by using a web-browsing history dataset in Section[?].
%In Section~\ref{sec: kanonysol} we investigate the $k$-anonymization technique as an additional privacy protection mechanism applied on histograms of users data.
In Section~\ref{sec: possiblesol}, we analyze the efficacy of our algorithm if additional privacy enhancing techniques, such as $k$-anonymization, are applied to histograms of users' data.
%We   experimentally evaluate our solution by using a Call Data Records dataset in Section[?] and by using a web-browsing history dataset in Section[?].
We conclude the paper with some discussions in Section~\ref{sec: conc}.

\section{Related work and contributions}\label{sec:relwork}
The user  matching studied in this paper is closely related to several problems that have been studied in other different communities.
In this section, we present a comparison of our approach with related problems from several areas, and highlight our contributions relative to existing work.

\subsection{Entity resolution}
%RRRR: This part has to be rewritten substantially. Are these really matching problems? If not, we must point out the differences! Look up ``Record linkage'' on Wikipedia! How about changing the title to ``Optimal Record Linkage by Matching Histograms''?
A matching problem studied in the database community is that of  identifying different data records that refer to the same real-world object~\cite{elmagarmid2007duplicate}. %CITE bhattacharya2007collective removed
Similarly, in natural-language processing,  the problem  of linking different mentions of the same underlying entity in text~\cite{bengtson2008understanding} is analogous to the objective in the user-matching problem. %CITE removed cai2010end rao2000can
Another example from the information-retrieval literature is the problem of classifying documents by their authors,  given documents from different authors with the same name~\cite{kalashnikov2008web}. %CITE removed qian2011combining
User matching  has also been studied in the social-networks community in which the objective is to identify different profiles that belong to the same underlying user~\cite{peled2013entity}.
Such problems fall under the umbrella-term \emph{entity resolution (ER)}~\cite{liu2013s}.
%The solution we propose in the present paper is aimed at the setting in which the only information available about the users is in the form of histograms.
In these problems, the available information about the users is often not in the form of histograms, and the solutions proposed are often based on heuristics and practical convenience; whereas the solution we propose in this paper  is specific to the setting in which the only information available about the users is in the form of histograms, and in this setting, the solution is optimal for minimizing the probability of misclassification.

%======RRRR=======

%what does the next line mean?

%=================

%Nevertheless, similarity measures are developed for identifying the pieces of information that belong to the same user~\cite{vosecky2009user}.
%\subsection{Literature on privacy}
\subsection{De-anonymization attacks}
Our work is also closely related to the literature on de-anonymization methods~\cite{Narayanan:2008:RDL:1397759.1398064},\cite{sweeney1997weaving} studied in the literature on privacy.
%Researchers from the fields of security and privacy have shown that linking records of data from different anonymized databases to publicly available information may expose sensitive private information of the users.
%For example, in~\cite{Narayanan:2008:RDL:1397759.1398064} the anonymized Netflix dataset is de-anonymized using user reviews from IMDB,  and in~\cite{sweeney1997weaving} medical records are de-anonymized with the help of external auxiliary information, namely ZIP code, birth date, and gender.
A number of works on de-anonymization focus on demonstrating  that even when users' data are anonymized, the data belonging to each user is often unique.
In such examples, an adversary who has access to auxiliary information about the users can de-anonymize the anonymized datasets by exploiting the uniqueness of the data belonging to each user.
For example, in~\cite{zang2011anonymization} the authors perform a study on the top $k$ locations most frequently visited by users in a nationwide call-data record (CDR) dataset.
They consider various levels of spatial granularity (such as sector, cell, zip code, city, state, and country) and temporal granularity (such as day and month), and they show that the most frequently visited locations can act as quasi-identifiers to re-identify anonymous users.
Thus an adversary can de-anonymize such a dataset by obtaining access to auxiliary information about the users' zip codes and times of activity.
The  adversary's goal is  essentially a matching task, i.e., the adversary seeks to match the auxiliary information about the users with the unique aspects of the users' data.
Some other works such as~\cite{xiao2010finding, ma2010privacy, freudiger2012evaluating, gambs2014anonymization, de2008identification, zang2011anonymization,machanavajjhala2008privacy} study the uniqueness of mobility data traces.
There is a line of work on studying the uniqueness of web browsing history patterns of users~\cite{olejnik2014uniqueness,yen2012host}.
In~\cite{olejnik2014uniqueness} the authors consider a dataset where every record is the set of visited websites by  a user during some period of time.
The authors investigate how unique is a single user's record compared to other users' records in the dataset.

Although our work is related to de-anonymization, it differs in several aspects.
%The user-matching problem we study  is closely related to these works on de-anonymization.
%However, our work differs in several respects.
First, we assume that the only information about the users in the two datasets are time-averaged statistics of the users' data.
In most works on user matching and de-anonymization~\cite{Narayanan:2008:RDL:1397759.1398064,sharad2013anonymizing,srivatsa2012deanonymizing}, the vulnerability to privacy breaches often arises due to the sparsity of the temporal evolution of the users' data.
For instance, the fact that a user watched and rated a movie during a particular time-period or  was at a  specific location during a particular time can be used to easily identify the user's data from the anonymized dataset.
Other de-anonymization works focus on identifying the temporal patterns of the data collected from the users.
For example, in~\cite{gambs2014anonymization, de2008identification}, a Markov model is constructed based on the temporal evolution of the mobility patterns of the users, and then similarity measures are used for de-anonymization.
Such temporal information in the users' data, however, is removed when only statistics in the form of histograms from each user is collected or released.
Often this results in a much lower uniqueness in the information available about the users; hence matching  users' statistics is, in general, much more difficult than matching users' datasets.

Second, we assume that the two sets of the statistical information are mutually statistically independent.
For example, in the case of mobility data, this could be because the two datasets were obtained over different time periods.
%We seek to perform the matching by only exploiting the fact that the statistics of the users' habits remain unchanged across the two datasets.
We seek to perform the matching by only exploiting the fact that users' habits remain stationary and ergodic across the two datasets.
This is in contrast to the approach of works such as~\cite{ma2010privacy,monhidverblo13,freudiger2012evaluating} that perform de-anonymization by using auxiliary information collected over the same period of time as the anonymized dataset.
In such cases, the auxiliary information is not independent of the anonymized user data.
In~\cite{olejnik2014uniqueness}, the authors investigate the stability of the set of visited websites by a user across time. 
In particular, they record the set  of visited websites by a user  during one day.
They use the Jaccard index to measure the similarity between the sets collected for one user over  different days.
They show that the set of visited websites by a user is stable during a four-week period.
A special case of our work is when the labeled and unlabeled histograms are obtained from the same source in different time periods.
The accuracy of our matching algorithm in such cases is dependent on how how much the statistical characteristics of the data is preserved over time.

Third, we perform simultaneous matching of the information available about all users and not one user at a time.
Simultaneous matching takes into account all  the information  available about the users at the same time, and hence outperforms matching users one at a time.
Simultaneously taking into account all the information for various attacks has already been employed in different domains~\cite{shokri2011quantifying,afroz2014doppelganger,danezis2009vida,troncoso2008perfect}
and in this paper we employ it in the domain of  histogram matching.

There is also a related line of work on graph de-anonymization,   also known as graph alignment~\cite{pedarsani2011privacy,sharad2013anonymizing,srivatsa2012deanonymizing}.
It is the problem of matching the nodes across  two similar graphs, where the only available information is the two graphs.
For example, given the graph of connections between users on two different social networks (e.g., Facebook and LinkedIn), it might be possible to match users across the two social networks by exploiting the fact that the structure of the underlying graphs are expected to be similar.
This problem is  different from that studied in the present paper because, in our setting, the graph-based connections among the users are not available.

\subsection{Supervised learning}
\begin{figure}
\centering
\subfigure[Histogram classification]{
\psfrag{S}[l][Bc]{\parbox[c]{1.4cm}{\begin{center} Training \\ histograms\end{center}}}
\psfrag{U}[r][Bc]{\parbox[c]{1.2cm}{\begin{center} Test \\ histogram \end{center}}}
\includegraphics[width=.45\columnwidth]{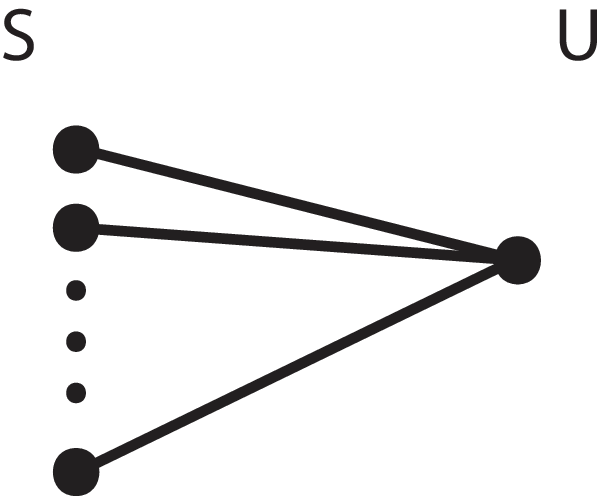}
\label{fig: histClassificationa}
}
\hfill
\subfigure[Histogram matching]{
\psfrag{S}[l][Bc]{\parbox[c]{1.4cm}{\begin{center} Training \\ histograms\end{center}}}
\psfrag{U}[r][Bc]{\parbox[c]{1.4cm}{\begin{center} Test \\ histograms \end{center}}}
\includegraphics[width=.45\columnwidth]{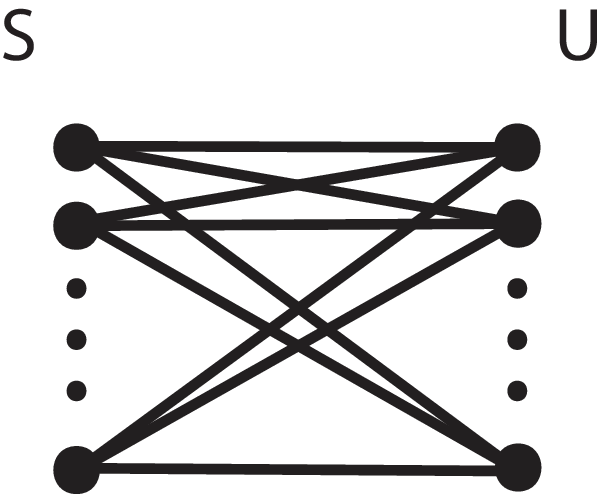}
\label{fig: histClassificationb}
}
\caption{(a) The problem of histogram classification, which is to to classify the test histogram to the correct class based on the training histograms.  (b) The problem of histogram matching studied in this paper, which is to simultaneously classify the test histograms to the training histograms subject to the constraint that each test histogram belongs to a distinct class.}
 \label{fig: histClassification}
\end{figure}

%In essence, this paper, we propose a new method for identifying a set of users by matching the statistical properties of their data with data collected in an independent experiment.
The matching task studied in this paper is closely related to the classification task studied in \emph{supervised learning}~\cite{hastibfri09}, where the objective is to classify test data to the correct class based on labeled training data observed under each of the classes.
Nevertheless, a key aspect of our approach  that differs from supervised learning is that we seek to simultaneously classify test data that belong to a group of users subject to the constraint that each user belongs to a distinct class (refer to Figure~\ref{fig: histClassification}).
Thus our solution, originally introduced in~\cite{unniMNainiAlert13, unn13}, can be interpreted as a solution to a \emph{constrained classification} problem.
Our solution is tailored to the setting in which the available information is in the form of histograms.
It could be possible to extend this solution to more general kinds of data by combining the matching algorithm presented in this work with feature extraction techniques in machine learning~\cite{hastibfri09}.
%=========RRRR ==============

%Add a figure here with two subfigures, one showing classification of histograms and another matching of histograms. Be consistent in presentation with Figure~\ref{fig: bipartitematch4traj1}.

%========RRRR more on related work=======
%
%One thing to look up is the following: In many applications, when data is released into the public domain, the matching of indices to users are periodically varied for enhancing privacy. E.g., the dataset we got from EPFL network admin. Is there some work that studies the issues with such anonymization? At least we must mention this as a specific application.
%
%==========RRRR================

\subsection{Contributions}\label{sec:contribution}
Compared to existing works on the user-matching problem, our work is unique in several respects.
Our main contributions can be summarized as follows:
\begin{itemize}
\item We demonstrate that statistics about users' behaviors contain a significant amount of information that can be used as \emph{fingerprints} to uniquely identify users, by an adversary who has access to auxiliary information about the users.
Moreover, we show that identification by using only data statistics can sometimes result in  accuracy higher than existing methods based on more complicated data models (e.g., Markov Chains). %[RRRR: should change]
\item We evaluate a provably optimal algorithm for matching users' statistics on three datasets of diverse nature and demonstrate that it outperforms heuristics-based methods.
We address the practical setting of performing the matching across distinct sets of users.
\item We compare the performance of our algorithm with different parameters and under different settings, such as user configuration and data resolution.
We verify that, in particular,  matching users \emph{simultaneously} leads to a matching accuracy significantly higher  than matching \emph{one user at a time}.
\item We analyze the performance of the matching algorithm under different privacy-preserving mechanisms such as data obfuscation and $k$-anonymization.
\end{itemize}

\section{Problem Statement and Proposed Solution}\label{sec: Ch4problemstatement}
We assume that the data belonging to each user in our system follows some fixed underlying probability law that is unknown a priori.
The probability law associated with each user is unique and captures the habits of the user.
%Let us assume we have $\pop$ users, where  each user's \emph{habit} is governed by a probability law.
For example, in the case of web-browsing histories, the probability law captures the user's relative preferences  for various websites.
Similarly, in the case of shopping data, the probability law could represent shopping preferences and, in the case of mobility data, the law could represent the preferences for visiting various locations.
%By habit of a user we mean, for example, the web-browsing habit (i.e., the kind of pages the user visits), the kind of items the user shops from Amazon, or the pattern in the user's visited locations (i.e., the probability model governing the mobility of the user).
In the basic version of the user-matching problem, we are given two datasets corresponding to the same set of users, and the task is to match users across the two datasets by exploiting the fact that the underlying probability law of each user is unique.
We will later generalize this to the setting in which the two datasets belong to different sets of users.
Throughout this paper, we focus on the specific setting in which each dataset reveals only the histograms of each user's data and not the data itself.
%We assume that we are given information about users' habits in the form of histograms by two datasets.
%Our task is to match the histograms that are associated with the same underlying user.
%We denote by \emph{adversary} the party that performs the user matching task.
We use the term \emph{adversary} to denote the entity that performs the user-matching task.
We use feminine pronouns for referring to the users and masculine pronoun for referring to the adversary.
In the following, we state the problem mathematically.

\subsection{Problem statement}\label{sec:probstmnt}
Consider a discrete alphabet set $\zstate=\{S_1, S_2, \ldots, S_\staten\}$ of size $|\zstate|=\staten$ and a  set of $\pop$ users labeled $1,2,\ldots, \pop$.
The set $\zstate$ represents the set of all possible values that can be taken by each instance of the data belonging to each user.
For example, in the case of web-browsing data, $\zstate$ is the set of all websites that a user could visit, and in the case of mobility data, $\zstate$ is the set of all possible locations (e.g., regions of a city) that a user could visit.
%[RRRR: spatial location not very clear. E.g., it cannot be GPS coordinates!. It represents locations with semantic meaning, such as restaurant, home, etc.]
For the purpose of illustration, in the rest of this section, we will focus on the example of mobility data.
%\notes{the dataset actually contains other two-week periods}
%
%==========RRRR Do we need this ===========
%
%Let  $\prsim(\zstate)$ denote the ($\staten-1$)--dimensional simplex of all probability distributions defined on $\zstate$:
%\begin{equation}
%\prsim(\zstate)=\left\{ \left[ p_1, p_2, \ldots, p_{\staten} \right] \Big\vert \forall l \; p_l \geq 0,  \sum_{l}^{\staten}p_l =1 \right\}.
%\end{equation}
%==================RRRR====================

For a data string $s=[s(1), s(2), \ldots, s(\timen)]  \in \zstate^\timen$ of length $\timen$, we use $\Gamma_{s}$ to denote the histogram (i.e., empirical distribution) of the string defined as
\begin{equation}\label{eq: defemperical}
\Gamma_{s}(l) = \frac{1}{\timen}\sum_{t=1}^\timen \clI\{s(t) = S_l\},\; l=1,2,\ldots,  \staten.
\end{equation}
In the simplest version of the user-matching problem studied in this paper, we are given two sets of histograms of the data generated by each of the users.
Let set $\SetSa = \{\Gamma_{x_1}, \Gamma_{x_2}, \ldots, \Gamma_{x_\pop}\}$ represent a set of \emph{unlabeled} histograms each generated by a distinct user, and let $\SetSb = \{\Gamma_{y_1}, \Gamma_{y_2}, \ldots, \Gamma_{y_\pop}\}$ represent a set of \emph{labeled} histograms each generated by a distinct user.
%Sets $S_1$ and $S_2$ can be interpreted as the training and the test sets, respectively.
Here  $\SetSa$ and $\SetSb$ represent the histograms contained in  two datasets.
In the case of mobility data, $\SetSa$ is a set of anonymized histograms of users' mobility traces that are released, and $\SetSb$  represents the  auxiliary histograms of the users' mobility traces, which is obtained by an adversary by tracking the users over a time period.
In other applications, the auxiliary histograms can be obtained by the adversary by using publicly available information.
In both cases, the adversary is aware of the users' identities in the second dataset, and seeks to decode the identities of the users in the first anonymized set of histograms.
%Here $\SetSa$ represents, for example, the unlabeled (i.e., anonymized) statistics of users' mobility traces that are released,  and $\SetSb$ represents the
%The information $\SetSb$ is assumed to be \emph{statistically independent} of $\SetSa$.
The histograms of each user are assumed to be statistically independent of those of others. 
Furthermore, for each user, the histogram generated by the user in the first dataset is assumed to be independent of the histogram in the second dataset.
In the mobility example, independence is a reasonable assumption provided that there is no overlap between the time-periods over which the histograms in $\SetSa$ and $\SetSb$ are computed.
For example, $\SetSa$  contains histograms collected over a week and $\SetSb$ contains histograms collected over the following week.

%The information in $\SetSb$ could be collected, for example, by tracking the users by a location based service provider who is acting as the adversary.

In the matching problem, the objective of the adversary is to determine the true matching between the histograms of $\psi_1$ and $\psi_2$.
We represent the ground truth via an unknown permutation function,
\begin{equation}\label{eq:Ch4perturb1}
\permU: \{1,2,\ldots, \pop\} \mapsto \{1,2,\ldots, \pop\}
\end{equation}
such that, in reality, for each $i \in \{1,2,\ldots,N\}$, the histograms $\Gamma_{x_{\permU(i)}}$ and $\Gamma_{y_i}$  are generated by the same user $i$.
The objective in the matching problem is, equivalently, to estimate $\permU$.
In Section~\ref{sec: uncbipartitematch1}, we discuss the practical setting where the histograms in the sets $\SetSa$ and $\SetSb$ are generated by different sets of users.
%In the remainder of the paper we use \emph{adversary} to denote the identity that performs the matching.

%denote the user who generated histograms $\Gamma_{x_{\permU(i)}}$ and $\Gamma_{y_i}$ where
%\begin{equation}\label{eq:Ch4perturb1}
%\permU: \{1,2,\ldots, \pop\} \mapsto \{1,2,\ldots, \pop\}
%\end{equation}
% is some unknown permutation.
%When $L=K$, the function $\pi$ is just some unknown permutation.
%Let $\prdisvec_i$ denote a probability measure on $\zstate$ that captures the probability law followed by data from source $i$.
%The user identification problem, in other words, the de-anonymization problem that the adversary needs to solve, is to match each histogram from set $\SetSb$ to the histogram from $\SetSa$ produced by the corresponding user.
%Equivalently, the objective is to  estimate $\permU$.
%The special case of this estimation problem when $\pop=1$ was studied by Gutman~\cite{gut89}.
%In the present paper, we study the other extreme case of $\pop$.

\subsection{Potential approach: Weighted bipartite matching}\label{sec: proposedapproach1}
\begin{figure}
\centering
\psfrag{x}[Bc][Bc]{\small{$\Gamma_{x_1}$}}
\psfrag{a}[Bc][Bc]{\small{$\Gamma_{x_2}$}}
\psfrag{b}[Bc][Bc]{\small{$\Gamma_{x_\pop}$}}
\psfrag{y}[Bc][Bc]{\small{$\Gamma_{y_1}$}}
\psfrag{c}[Bc][Bc]{\small{$\Gamma_{y_2}$}}
\psfrag{d}[Bc][Bc]{\small{$\Gamma_{y_\pop}$}}
\psfrag{w}[Bc][Bc]{\small{$w_{11}$}}
\psfrag{S}[Bc][l]{\parbox[c]{3cm}{\begin{center} Dataset one \\ histograms ($\SetSa$)\end{center}}}
\psfrag{U}[Bc][l]{\parbox[c]{3cm}{\begin{center} Dataset two \\ histograms ($\SetSb$)\end{center}}}
\includegraphics[width=.5\columnwidth]{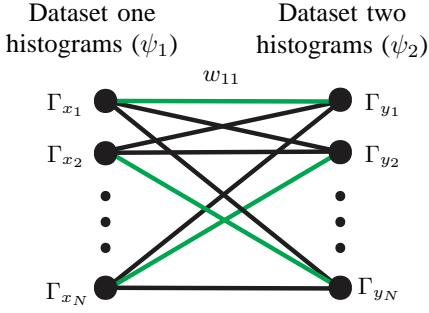}
%\label{fig: bipartitematch4traj1b}
\caption{The problem of matching histograms across two sets can be  visualized as a matching problem on a weighted bipartite graph. Corresponding to the $\pop!$ different permutations, there are $\pop!$ possible maximal matchings on $G$. The green edges represents the correct matching associated with $\permU$ in~\eqref{eq:Ch4perturb1}.
The solution can be obtained via a weighted bipartite matching algorithm on the graph with appropriate edge weights.}
%The hypothesis testing task thus is equivalent to identifying the correct maximal matching.
 \label{fig: bipartitematch4traj1}
\end{figure}

The problem of matching histograms across two sets can be best visualized as a matching problem on a bipartite graph.
Let $G = (V_1, V_2, E)$ be a complete bipartite graph where each vertex in the set $V_1$ (respectively, set $V_2$) is associated with  a unique histogram in the set $\SetSa$ (respectively, set $\SetSb$).
%Let $G = (V,E)$ denote a complete bipartite graph where each vertex in the set $V$ of vertices is associated with  a unique element in $S_1 \cup S_2$.
There exists an edge from each element  in $V_1$ to each element in $V_2$ and no edges between elements in $V_1$ or $V_2$.
%There exists an edge from each element $i$ in $S_1$ to each element $j$ in $S_2$ and no edges between elements in $S_1$ or $S_2$.
Hence we have a complete bipartite graph where $V_1$ and $V_2$ form the two parts.
%Hence we have a complete bipartite graph where $S_1$ and $S_2$ form the two parts.
Let node $\z$ in set $V_1$ and node $i$ in set $V_2$ be associated with histogram $\Gamma_{x_\z}$ in $\SetSa$ and  $\Gamma_{y_i}$ in $\SetSb$, respectively.
The graph $G$ is illustrated in Figure~\ref{fig: bipartitematch4traj1}.

A \emph{matching} in graph $G$ is a subset of edges $E$ of $G$ such that no two edges  in the subset share a vertex. %violate
A \emph{maximal matching} is a matching  such that the addition of any edge to the subset violates the matching property.
Let $\permU_\j$ be a permutation of $\{1,2,\ldots,\pop\}$, for $\j=1,2,\ldots, \pop!$.
There are $\pop!$ possible {maximal matchings} on $G$ corresponding to the $\pop!$ different permutations. 
The matching corresponding to permutation $\sigma_\j$ is the  matching in which each node $i$ from set $V_2$  is mapped to node $\permU_\j(i)$ in $V_1$; in other words, histogram $\Gamma_{y_i}$ in $\SetSb$ is mapped to histogram $\Gamma_{x_{\permU_\j(i)}}$ in  $\SetSa$.
%A matching in graph $G$  in Figure~\ref{fig: bipartitematch4traj1} is shown by green edges.
The  matching associated with $\permU$ in~\eqref{eq:Ch4perturb1}  is shown by green edges in Figure~\ref{fig: bipartitematch4traj1}.
%The green edges represents the correct matching associated with $\permU$ in~\eqref{eq:Ch4perturb1}.

An intuitive approach for estimating the correct matching between the histograms is as follows.
Define a weight for every edge in $G$ such that the weight of the edge $w_{\z i}$ from $\z$ to $i$ is equal to some appropriately defined distance between the histograms $\Gamma_{x_\z}$ and $\Gamma_{y_i}$, i.e.,
\begin{equation}\label{eq: weightmetric}
w_{\z i} = \mbox{d}(\Gamma_{x_\z}, \Gamma_{y_i})
\end{equation}
for some distance measure $ \mbox{d}(.)$.
Now perform a minimum-weight maximal bipartite matching on the resultant weighted bipartite graph.
The minimum-weight maximal matching corresponds to a configuration where the sum of the distances between the matched histograms is minimum,  hence  expected to provide a good estimate for the correct matching.

%A standard  approach in finding a matching between the histograms of $\SetSa$ and $\SetSb$ is to define a \emph{similarity} measure (or, equivalently, a \emph{distance} measure) between each pair of histograms from  $\SetSa$ and  $\SetSb$ and seek the matching that yields the maximum    summation of the similarities (or, equivalently, the minimum summation of distances) between the matched pairs.
%Finding such a matching is equivalent to solving a  weighted bipartite matching problem on the graph $G$,  where to each edge $e_{zi}$ corresponds a weight $w_{zi}$ which represents the similarity (or distance) between histograms $\Gamma_{x_z}$  and $\Gamma_{y_i}$.

The relevant questions that arise here are: What is a good choice for the distance measure between histograms and does the choice of measure depend on the nature of the data or can there be a general-purpose measure?
The literature contains various choices of prevalent distance measures that can be used in the weight function.
For example, in~\cite{zang2011anonymization} the authors  use the \emph{cosine} distance between the histograms of the number of calls of users at different GSM antennas as a distance measure for analyzing the call behavior of users.
The cosine distance between histograms $\Gamma_{x_\z}$ and $\Gamma_{y_i}$ is defines as
\begin{equation}\label{eq: Ch4cossim1}
%w_{zi}^{\mbox{cos}} = 1-\frac{w_{zi}^{\mbox{dot}}}{\norm{\Gamma_{x_z}}_2 \norm{\Gamma_{y_i}}_2},
w_{\z i}^{\mbox{cos}} = 1-\frac{ \left<\Gamma_{x_\z} , \Gamma_{y_i}\right> }{\norm{\Gamma_{x_\z}}_2 \norm{\Gamma_{y_i}}_2},
\end{equation}
where $\left<\Gamma_{x_\z} , \Gamma_{y_i}\right>$ is the dot product between the histograms, which we denote by $w_{\z i}^{\mbox{dot}}$,
\begin{equation}\label{eq: Ch4dotsim1}
w_{\z i}^{\mbox{dot}} =\left<\Gamma_{x_\z} , \Gamma_{y_i}\right> = \sum_{l=1}^{\staten}  \Gamma_{x_\z}(l)\Gamma_{y_i}(l),
\end{equation}
and $\norm{\Gamma_{s}}_2 = \sqrt{\sum_{l=1}^\staten {\Gamma_{s}(l)}^2}$.
Another heuristic measure often used in the machine learning community is the \emph{$l_1$ distance} between the histograms, given by
\begin{equation}\label{eq: Ch4lpdist1}
w_{\z i}^{l_1} = \norm{\Gamma_{x_\z} - \Gamma_{y_i}}_1= \sum_{l=1}^{\staten}  \left|\Gamma_{x_\z}(l)-\Gamma_{y_i}(l)\right|.
\end{equation}
Alternatively, we can use a similarity measure, such as the \emph{dot product} defined in~\eqref{eq: Ch4dotsim1} as the weight function in~\eqref{eq: weightmetric}.
We then identify the best permutation by using a \emph{maximum} weight matching on the resultant weighted bipartite graph.
In the next subsection, we present a new choice of the weight function and argue that it is a judicious choice.

\subsection{Optimal solution via hypothesis testing interpretation}\label{sec: hyptest}
The problem of finding the matching between the histograms of $\SetSa$ and $\SetSb$   can  be viewed as a multi-hypothesis testing problem with $\pop!$ hypotheses, $\{H_1,H_2,\ldots,H_{N!}\}$, where hypothesis $H_\j$ corresponds to permutation $\sigma_\j$, for $\j=1,2,\ldots, \pop!$.
In the hypothesis testing framework, we study decision rules by using probability of error under the different hypotheses as the performance metric.
Typical solutions to hypothesis testing problems seek the decision rule that leads to an optimal trade off between various error probabilities under the different hypotheses.
In our prior works~\cite{unniMNainiAlert13, unn13}, we  showed that, when each user's data is generated by an i.i.d. process governed by her probability law, an optimal trade-off between the various error probabilities for the matching problem is obtained by deciding in favor of the hypothesis corresponding to the minimum-weight maximal matching on the bipartite graph $G$ with edge weights
\begin{equation}\label{eqn:weights}
w_{\z i} = D(\Gamma_{x_\z} \| \half({\Gamma_{x_\z} + \Gamma_{y_i}}) )+ D(\Gamma_{y_i} \| \half({\Gamma_{x_\z} + \Gamma_{y_i}})).
\end{equation}
In~\eqref{eqn:weights}, $D(\cdot\|\cdot)$ is the Kullback-Leibler divergence function~\cite{covtho06}, defined as
$$D(\prdis\|\mu)=\sum_{l=1}^{\staten}\prdis(l)\log\left(\prdis(l)/\mu(l)\right).$$
The weight $w_{\z i}$ in~\eqref{eqn:weights} satisfies $0 \leq w_{\z i} \leq 2\log(2)$ and it is equal to $0$ when  $\Gamma_{x_\z}=\Gamma_{y_i}$ and equal to $2\log(2)$ when the histograms have disjoint support (i.e., when $\nexists\; l$ such that $\Gamma_{x_\z}(l),  \Gamma_{y_i}(l) >0$).
The exact optimality result is based on an asymptotic analysis of error probabilities as the length $T$ of the data strings increases to infinity.
It is shown in~\cite{unniMNainiAlert13, unn13} that a variant of this test yields optimal trade-offs between the \emph{error exponents} under the different hypotheses.
%\notes{emphasis that its our results}
For the sake of completeness, we provide the intuition behind the choice of the metric~\eqref{eqn:weights} in the following setting.
To every user $i$, we associate a probability distribution $\prdisvec_i \in \prsim_{\staten-1}$.
The probability distributions are distinct, which means that $\prdisvec_i \neq \prdisvec_\j$  for $i \neq \j$, but they are unknown.
Suppose that each user $i$ generates data in an i.i.d. manner  from the distribution $\prdisvec_i$.
Consider a set $\SetSa = \{x_1, x_2, \ldots, x_\pop\}$ of \emph{unlabeled} strings of length $\timen$ each generated by a distinct user, and an independent set $\SetSb = \{y_1, y_2, \ldots, y_\pop\}$ of \emph{labeled} strings of length $\timen$ each generated by a distinct user.
Let $i$ denote the user who generated strings $x_{\permU(i)} \in \zstate^\timen$ and $y_i\in \zstate^\timen$ where $\permU$ is given in~\eqref{eq:Ch4perturb1}.

A commonly used solution for multi-hypothesis testing problems is to identify the maximum-likelihood (ML) hypothesis, which is the hypothesis under which the log-likelihood of the observations is maximized.
In our problem, however, the underlying probability distributions $\prdisvec_i$'s of the users are unknown, thus the log-likelihood has to be replaced with the \emph{generalized} log-likelihood.
The first step is therefore to compute the generalized log-likelihood.
For hypothesis $H_\j$ the generalized likelihood is obtained by maximizing the likelihood function over all possible choices of the $\prdisvec_i$'s, and is given by
\begin{align}\label{eq:CH4Hypexp1}
%\llikeli(H_j) ={\sup_{\prdisvec_1,\prdisvec_2,\ldots,\prdisvec_\pop} \sum_{i=1}^\pop \left[\log \prdisvec_i(x_{\bm{\sigma}_j(i)}) + \log \prdisvec_i(y_i)\right]}.
\llikeli(H_\j) =\sup_{\prdisvec_1,\prdisvec_2,\ldots,\prdisvec_\pop} \sum_{i=1}^\pop \sum_{t=1}^\timen \Big[&\log\left( \prdis_i(x_{\permU_\j(i)}(t)) \right) \nonumber \\ &+ \log\left(\prdis_i(y_i(t))\right)\Big].
\end{align}
%We remark that in~\eqref{eq:CH4Hypexp1}, we take  a supremum over all probability distributions $\prdisvec_i$s, because the underlying probability distributions are unknown.
%We remark that because the underlying probability distributions $\prdisvec_i$s are unknown, the log-likelihood of the observations is computed by taking a supremum over all probability distributions in~\eqref{eq:CH4Hypexp1}.
%The computed log-likelihood  in~\eqref{eq:CH4Hypexp1} is thus called the generalized log-likelihood.
It is known that for an i.i.d.-generated string, the maximum likelihood estimator of the underlying distribution is given by the empirical distribution of the string.
Hence, it is easy to see that each of the $\pop$ terms in the summation~\eqref{eq:CH4Hypexp1} is maximized by setting $\prdisvec_i = \half({\Gamma_{x_{\permU_\j(i)}} + \Gamma_{y_i}})$,
for $i=1,2,\ldots,\pop$.
We can therefore rewrite~\eqref{eq:CH4Hypexp1} as
\begin{align}\label{eq:CH4Hypexp2}
&\llikeli(H_\j) = -2\timen \sum_{i=1}^\pop\Big\{\clH(\Gamma_{x_{\permU_\j(i)}}) + \clH(\Gamma_{y_i})   \nonumber \\ &\!+\!D(\Gamma_{x_{\permU_\j(i)}} \| \half({\Gamma_{x_{\permU_\j(i)}}\!+\! \Gamma_{y_i}}) )\! +\! D(\Gamma_{y_i} \| \half({\Gamma_{x_{\permU_\j(i)}}\!+\! \Gamma_{y_i}})) \Big\},
\end{align}
%\begin{align}\label{eq:CH4Hypexp2}
%&\llikeli(H_\j) = -2\timen \sum_{i=1}^\pop\Big\{\clH(\Gamma_{x_{\permU_\j(i)}}) + \clH(\Gamma_{y_i})   \nonumber \\ & %+D\left(\Gamma_{x_{\permU_\j(i)}} \| \frac{\Gamma_{x_{\permU_\j(i)}} + \Gamma_{y_i}}{2} \right) + D\left(\Gamma_{y_i} \| %\frac{\Gamma_{x_{\permU_\j(i)}} + \Gamma_{y_i}}{2}\right) \Big\},
%\end{align}
where $\clH(.)$ is the Shannon entropy function~\cite{covtho06}, defined as $\clH(\prdisvec)=-\sum_{l=1}^{\staten}\prdis(l)\log\left(\prdis(l)\right)$.

The maximum generalized likelihood  solution is given by
$ \what H = \argmax_{H_\j} \llikeli(H_\j)$.
Given the sets of histograms $\{\Gamma_{x_\z}\}$ and $\{\Gamma_{y_i}\}$, the term $\sum_{i=1}^{\pop} \clH(\Gamma_{x_{\permU_\j(i)}}) +\clH(\Gamma_{y_i})$ in~\eqref{eq:CH4Hypexp2} is a constant term that does not depend on the hypothesis $H_\j$.
Hence, by removing this constant term we can show that $\what H = \argmin_{H_\j} D(H_\j)$, where $D(H_\j) = \sum_{i=1}^\pop w_{{\permU_\j(i)}{i}}$ with $w_{\z i}$ given in~\eqref{eqn:weights}.
Hence, $\what H$ can be interpreted as the hypothesis corresponding to the minimum-weight maximal matching on the complete bipartite graph $G$ in Figure~\ref{fig: bipartitematch4traj1} with weights~\eqref{eqn:weights}.

%\notes{Note that technically we need a min weight "maximal" matching}

%%%%%%%%%

Although this optimality result was established for i.i.d. processes, we argue that the solution is a reasonable approach to use for the matching problem, provided that each user's habits follow a probability law that is stationary and ergodic.
In such cases, we expect the histograms of each user in the two datasets to be similar, hence the solution for i.i.d. data is well-justified.
%We argue that  the distance measure given in~\eqref{eqn:weights} for the i.i.d. setting is also justified in a practical scenario where the probability laws governing users' habits remain the same across the two  datasets.
Therefore, in this paper, we propose to use the solution given by the minimum-weight maximal matching on $G$ with the weight metric in~\eqref{eqn:weights}.
% in such a practical setting where the probability laws governing users' habits remain the same across the two  datasets, and where the only available information about the users are in the form of histograms.
We demonstrate, in our experiments in Section~\ref{sec: experiments}, that the matching accuracy obtained by using~\eqref{eqn:weights} is indeed higher than those obtained by using~\eqref{eq: Ch4cossim1},~\eqref{eq: Ch4dotsim1}, and~\eqref{eq: Ch4lpdist1} under various settings.

\subsection{Generalization to different sets of distinct users}\label{sec: uncbipartitematch1}
%[TBD: Here we address the case where we have different sets of distinct users and we have an upper bound on the number of common users.]
Denote by $\USetSa$  the set of users who generate the histograms  $\SetSa$ and by $\USetSb$  the set of users who generate the histograms  $\SetSb$.
So far, we have assumed that the two sets of histograms $\SetSa$ and $\SetSb$ are generated by the same set of $\pop$ users; i.e.,  $\USetSa=\USetSb$ with $|\USetSa|=\pop$.
In practice, however, the histograms in sets $\SetSa$ and $\SetSb$  can belong to  different sets of distinct users, that is,  $\USetSa\neq \USetSb$.
%For example, the histograms   $\SetSb$  collected by the adversary might belong to a much larger set of users compared to the histograms  $\SetSa$.
When $\USetSa\neq \USetSb$, the  adversary needs to solve  the matching problem of identifying the set $\USetSa \cap \USetSb$ and of identifying the matching between the labeled and the unlabeled histograms belonging to the users in the set $\USetSa \cap \USetSb$.
Our matching solution and optimality result can be extended to the case $\USetSa\neq \USetSb$~\cite{unn13}.
%In this section, we generalize our matching approach to the  case $\USetSa\neq \USetSb$.
Let the number of users in sets  $\USetSa$ and $\USetSb$ (i.e, number of histograms in sets $\SetSa$ and $\SetSb$) be $\pop$ and $\pop^\prime$, respectively.
We assume that the probability law of every user in the set $\USetSa \cup \USetSb$ is distinct.
Without  loss of generality, we assume that $\pop^\prime > \pop$, i.e., there are more labeled histograms than unlabeled histograms.

\begin{figure}
\centering
\subfigure[$\USetSa \subset \USetSb$]{
\psfrag{a}[Bc][Bc]{\small{$\Gamma_{x_1}$}}
\psfrag{e}[Bc][Bc]{\small{$\Gamma_{x_2}$}}
\psfrag{b}[Bc][Bc]{\small{$\Gamma_{x_\pop}$}}
\psfrag{c}[Bc][Bc]{\small{$\Gamma_{y_1}$}}
\psfrag{f}[Bc][Bc]{\small{$\Gamma_{y_2}$}}
%\psfrag{d}[Bc][Bc]{} %{\small{$\Gamma_{y_{\pop^\prime}}$}}
\psfrag{d}[Bc][Bc]{\small{$\Gamma_{y_{\pop^\prime}}$}}
\psfrag{w}[Bc][Bc]{\small{$w_{11}$}}
%\psfrag{S}[Bc][Bc]{\small{Histograms $\SetSa$}}
%\psfrag{U}[Bc][Bc]{\small{Histograms $\SetSb$}}
\psfrag{S}[Bc][Bc]{\parbox[c]{1.3cm}{\small{\begin{center} Histograms \\ $\SetSa$\end{center}}}}
\psfrag{U}[Bc][Bc]{\parbox[c]{1.3cm}{\small{\begin{center} Histograms \\ $\SetSb$\end{center}}}}
\includegraphics[width=.45\columnwidth]{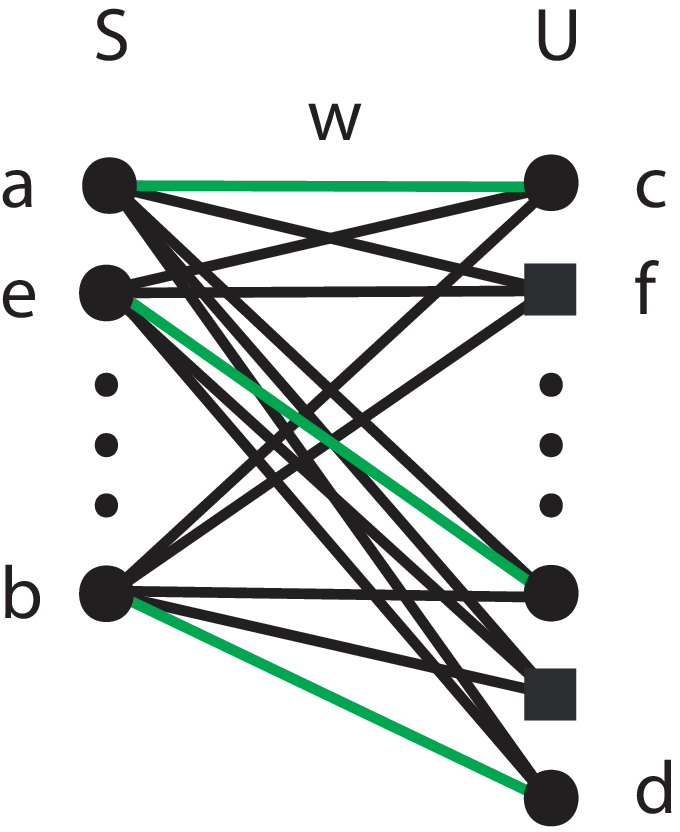}
\label{fig: bipartitematchsubset}
}
\hfill
\subfigure[$|\USetSa \cap \USetSb| = r < \pop$]{
\psfrag{a}[Bc][Bc]{\small{$\Gamma_{x_1}$}}
\psfrag{e}[Bc][Bc]{\small{$\Gamma_{x_2}$}}
\psfrag{b}[Bc][Bc]{\small{$\Gamma_{x_\pop}$}}
\psfrag{c}[Bc][Bc]{\small{$\Gamma_{y_1}$}}
\psfrag{f}[Bc][Bc]{\small{$\Gamma_{y_2}$}}
\psfrag{d}[Bc][Bc]{\small{$\Gamma_{y_{\pop^\prime}}$}}
\psfrag{w}[Bc][Bc]{\small{$w_{11}$}}
%\psfrag{S}[Bc][Bc]{\small{Histograms $\SetSa$}}
%\psfrag{U}[Bc][Bc]{\small{Histograms $\SetSb$}}
\psfrag{S}[Bc][Bc]{\parbox[c]{1.3cm}{\small{\begin{center} Histograms \\ $\SetSa$\end{center}}}}
\psfrag{U}[Bc][Bc]{\parbox[c]{1.3cm}{\small{\begin{center} Histograms \\ $\SetSb$\end{center}}}}
\includegraphics[width=.45\columnwidth]{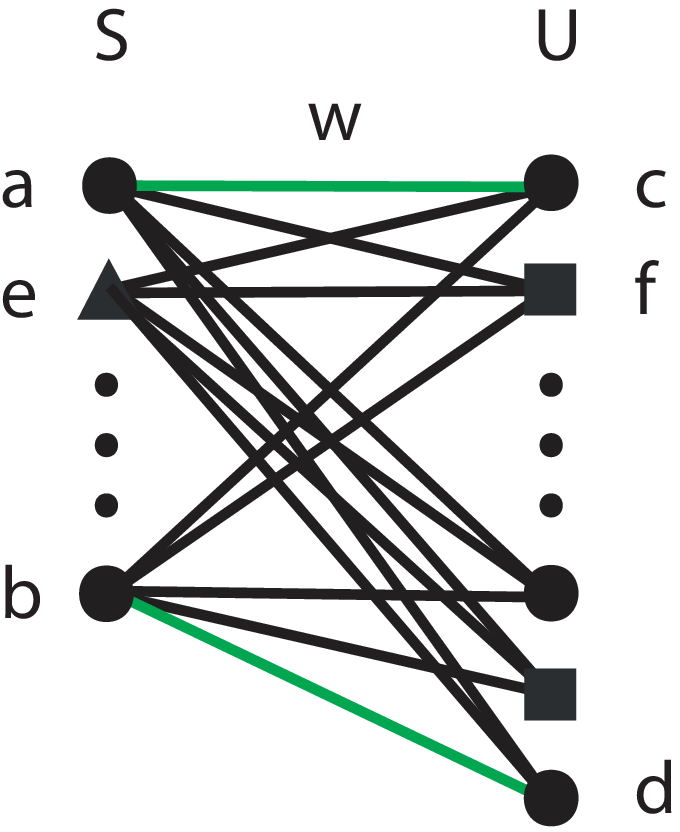}
\label{fig: bipartitematchunc1}
}
\caption{Matching problem when the histograms in sets $\SetSa$ and $\SetSb$  belong to  different sets of distinct users (i.e.,  $\USetSa \neq \USetSb$).
%Then, the matching problem that the adversary needs to solve is the problem of identifying the set $\USetSa \cap \USetSb$ and of identifying the matching between the labeled and the unlabeled histograms belonging to the users in the set $\USetSa \cap \USetSb$.
%For illustration purposes,
Histograms belonging to users in  $\USetSa \cap \USetSb$ are marked by black circles and histograms in the sets $\USetSa \backslash \USetSb$ and $\USetSb \backslash  \USetSa $ are marked by black triangles and squares, respectively.
In (a),  $\USetSa \subset \USetSb$ with $|\USetSa|=\pop$.
The proposed solution is given by the minimum-weight \emph{maximal} matching of the graph.
%In this case, a matching between all the $\pop$ labeled histograms and $\pop$ out of $\pop^\prime$ of the  unlabeled histograms can be obtained from the \emph{maximal} matching of the graph.
In (b), $|\USetSa \cap \USetSb|=r <\pop$.
%One approach for the adversary is to match all the $\pop$ unlabeled histograms to the labeled histograms, yielding at most $r$ correct matches and at least $\pop-r$ incorrect matches.
%The adversary can match only $r$ of the unlabeled histograms to the labeled histograms such that the  summation of the distance between the matched pairs  is minimized.
In this case, the proposed solution  is given by the minimum-weight matching with cardinality $r$ on the graph.
The green edges represent the correct matching between the histograms in the set $\USetSa \cap \USetSb$.}
%\label{fig: bipartitematchunc1}
\end{figure}

First, consider the case $\USetSa \subset \USetSb$.
Here $|\USetSa \cap \USetSb|=\pop$.
It represents the scenario where for each unlabeled histogram in set $\SetSa$ there exists  an associated labeled histogram in set $\SetSb$ that is generated by the same user, but not vice versa.
As before we construct the complete bipartite graph  $G = (V_1, V_2, E)$, where $|V_1|=\pop$, $|V_2|=\pop^\prime$, and edge weights are as in~\eqref{eqn:weights}.
%Consider the complete bipartite graph  $G = (V_1, V_2, E)$, where every node $z$ in $V_1$ (respectively, node $i$ in $V_2$) is associated with the unlabeled histogram $\Gamma_{x_z}$ (respectively, labeled histogram $\Gamma_{y_i}$)  and has corresponding edge weight  $w_{zi}$ given in~\eqref{eqn:weights}.
The graph is illustrated in Figure~\ref{fig: bipartitematchsubset}.
A matching between all the $\pop$ unlabeled histograms and $\pop$ out of $\pop^\prime$ of the  labeled histograms is a  \emph{maximal} matching on the graph  $G$.
Similarly to the case where  $\USetSa=\USetSb$, the optimal solution is given by the minimum-weight maximal matching in the graph $G$.
%   illustrated in Figure~\ref{fig: bipartitematch4traj1}, the Hungarian algorithm can be used to obtained the maximal matching, which has size $\pop$, in graph $G$ in $\bigO\left({\pop^\prime}^3\right)$ steps.

Now consider the more general case where $|\USetSa \cap \USetSb|=r < \pop$.
Furthermore, let the value of $r$ be known to the adversary.
This case represents the scenario where some labeled histograms in set $\SetSb$ are not associated with any unlabeled histograms in $\SetSa$ and vice versa.
%Furthermore, let $|\USetSa \cap \USetSb|=r$ and let the value of $r$ be known to the adversary.
%The matching problem that the adversary needs to solve is the problem of identifying the set $\USetSa \cap \USetSb$ and of identifying the matching between the labeled and the unlabeled histograms belonging to the users in the set $\USetSa \cap \USetSb$.
The graph $G$ for this case is illustrated in Figure~\ref{fig: bipartitematchunc1}.
%%%%%%%%%%%%%%%%%%%%%%%%%%%%%%%%%%%%%%%%%%%%%%%%%%%%%%%%%%%%%%%%%%%%%
If the adversary knows the value of $r$, he can try to match only  a set of $r$ unlabeled histograms to the labeled histograms such that the two sets  are \emph{as close as possible} to each other.
In other words, the adversary can try to choose  $r$ out of $\pop$ unlabeled histograms and match them to $r$ out of $\pop^\prime$ labeled histograms such that the summation of the distances between the matched pairs (given in~\eqref{eqn:weights}) is minimized.
Such a matching can be obtained from  a minimum-weight  matching \emph{with cardinality $r$}  on the graph $G$.
This problem is also known as the \emph{minimum-cost imperfect matching}~\cite{ramshaw2012weight}.
We experimentally evaluate this approach in Section~\ref{sec: expsingleton1}.
If the adversary does not know the value of $r$, he can still try to match $\min \{N,N'\}$ users.
However this leads to a larger fraction of incorrect matches, as we demonstrate in the experiments of Section~\ref{sec: expsingleton1}.

\subsection{Algorithms and complexity}%\label{sec: includenumcalls}
An important practical aspect of the user-matching task is the algorithm for obtaining the matching solution on the weighted graph $G$ and the associated time-complexity.
In Sections~\ref{sec: proposedapproach1} and~\ref{sec: uncbipartitematch1}, we discussed three different settings for finding the matching solution between $\SetSa$ and  $\SetSb$ on the graph $G$: (i) case $\USetSa = \USetSb$ depicted in Figure~\ref{fig: bipartitematch4traj1}; (ii) case $\USetSa \subset \USetSb$ depicted in Figure~\ref{fig: bipartitematchsubset}; and (iii)   the case $|\USetSa \cap \USetSb| = r < \pop$ depicted in Figure~\ref{fig: bipartitematchunc1}.
%In (i), (ii), and (iii), the matching solution can be obtained via a minimum-weight maximal matching on the graph $G$, and in  (iv) it can be obtained via
We require two kinds of algorithms to identify the solutions:
\begin{itemize}
\item[(A1)] Algorithm for identifying the minimum-weight maximal matching on $G$,
\item[(A2)] Algorithm for identifying the minimum-weight matching with a fixed cardinality $r$ on $G$.
\end{itemize}
The matching solution on $G$ can be obtained via (A1) in cases (i) and (ii), and via (A2) in case (iii).
We note that in case of using a similarity measure such as~\eqref{eq: Ch4dotsim1} as the choice of the weight function in $G$, the matching solution is identified via the maximum-weight matching on $G$.
In this case, after negating all the  edge-weight values and shifting them to make them positive, (A1) and (A2) can be used to identify the matching solution.

The Hungarian algorithm~\cite{fredman1987fibonacci} is a popular and efficient algorithm for (A1) and  can be adapted to solve (A2) as explained in~\cite{ramshaw2012weight}.
In our experiments, we use the Hungarian algorithm for (A1) and a polynomial-time algorithm, based on the theory of matroids (see, e.g.,~\cite[Ch. 8]{coocunpulsch11}), for (A2).
The time-complexity of obtaining the matching solution on the graph $G$ by using the Hungarian algorithm is $\bigO(|\USetSa| |\USetSb| |\USetSa \cap \USetSb|)$; i.e., it is $\bigO(\pop^3)$,   $\bigO(\pop^2 \pop^\prime)$, and $\bigO(\pop \pop^\prime r)$ for (i), (ii), and (iii), respectively.
In practice, the complexity can often be reduced significantly.
For instance, when histograms $\Gamma_{x_\z}$ and $\Gamma_{y_i}$ have disjoint support, then $w_{\z i}$ in~\eqref{eqn:weights} takes its maximum value, which is $2\log(2)$.
Then the edge connecting the corresponding vertices in $G$ can be removed, as it will almost certainly not be selected in the minimum-weight maximal matching.
If the resulting graph has $\mathcal{E}$  edges, then the complexity  is $\bigO(\mathcal{E} |\USetSa \cap\USetSb|)$.

In a practical implementation of this de-anonymization approach, the overall complexity depends on both the complexity of computing the edge weights in graph $G$ and of running the matching algorithm (A1) or (A2) on graph $G$. 
The former has complexity $\bigO(\pop \pop^\prime\staten)$ where $\staten$ is the number of locations.
In Section~\ref{sec: expruntim1} we present detailed time-complexity results of our de-anonymization approach.

An alternative approach for solving (A1) and (A2) is to use an approximate minimum-weight  matching algorithm on graph $G$ instead of the Hungarian algorithm.
Although finding the exact minimum-weight matching solution has the advantage of obtaining the maximum matching accuracy,  it brings the inherent computational complexity of weighted bipartite matching into our solution. 
This could hinder the applicability of our solution to very large datasets as the number of histograms becomes very large.
An alternative approach in dealing with very large datasets is to obtain an \emph{approximate} minimum-weight matching solution on graph $G$~\cite{Duan:2014:LAM:2578041.2529989}.
Although this approach reduces the matching accuracy, it makes it possible to find an approximate solution in reasonable time.
For example by using the approach in~\cite{Duan:2014:LAM:2578041.2529989}, a $(1-\epsilon)$-approximate matching solution to (A1) in case (i) can be obtained with complexity    $\bigO(\pop^2 \epsilon^{-1} \log \epsilon^{-1})$ instead of $\bigO(\pop^3)$.

\section{Experimental Evaluations}\label{sec: experiments}
In this section we compare the performance of the proposed matching algorithm with other methods for user identification.
Although numerous identification algorithms exist in the literature, we perform comparisons primarily with identification methods that rely only on histogram information as the focus of this paper is on such methods. Nevertheless, in  Section~\ref{sec: geolifecomprelated} we compare our approach with an existing Markov-based method, for which histograms are only a subset of the information available to the method. We show that by using only histograms we can still get better de-anonymization accuracy than the Markov-based approach that exploits more information from the dataset.

We test our matching algorithms on three datasets of different nature.
%In this section, we test our matching algorithms  on three datasets of different nature.
The first is a call-data records dataset, the second is a web browsing-history dataset, and the third is a dataset of GPS mobility traces.
In our experiments, a \emph{location} represents the coverage region of a GSM antenna, a website, and a region on the map in the first, second, and  third dataset, respectively.
We interpret the sequence of locations visited by a user as a data string.
Thus a user's histogram is simply the relative fractions of visits of the user to the different locations, within the time period considered.
For each dataset, we compute the histograms of the users over two different non-overlapping time periods to obtain the sets $\SetSa$ and $\SetSb$ described in Section~\ref{sec:probstmnt}.
We then construct the complete bipartite graphs $G$ shown in Figures~\ref{fig: bipartitematch4traj1},~\ref{fig: bipartitematchsubset}, and~\ref{fig: bipartitematchunc1} and apply the matching algorithms proposed in Sections~\ref{sec: Ch4problemstatement} and~\ref{sec: uncbipartitematch1} on this graph with appropriately chosen edge-weights.
We estimate the matching accuracy obtained with the different algorithms by calculating the percentage of common users (i.e., users in the set $\USetSa \cap \USetSb$) whose histograms are correctly matched.
We recall that we focus on the privacy from the perspective of the adversary and not of the users; hence, this particular choice for  notion of accuracy is reasonable.
%
%The focus of this section is to perform user matching by using only histogram information and to show that in such scenarios the choice of the metric is very important. 
%Therefore, we have  focused mainly on comparing our proposed metric with other metrics.
%Many of the other de-anonymization approaches in the literature process  other kinds of  information (e.g., timing information), hence it is not fair to compare our approach with them. 
%Nevertheless, in Section~\ref{sec: geolifematchingexp1} we compare our approach with an existing Markov-based method, for which histograms are only a subset of the information available to the method.
%We show that  by using only histograms we can still get better de-anonymization accuracy than the Markov-based approach that exploits more information from the dataset.

\subsection{Experiments  on call-data records (CDR)}\label{sec: CDRdataset}
%In this section we perform experiments on the Orange Call Data Records dataset~\cite{blondel2012data}.
%This dataset  consists of anonymized records of phone calls  between Orange customers (i.e., users)  in Ivory Coast~\cite{blondel2012data}.
%In order to protect the users' privacy, the dataset reflects different trade-offs in terms of the location's accuracy and the time span over which the trace is provided.
%In our experiments, the sets  $\SetSa$ and $\SetSb$ correspond to the histograms of the number of calls of users from different GSM cell towers (i.e., antennas) measured over two different non-overlapping time periods.
%Using \emph{only} the histogram of calls  of the various users from various locations in each time-period, we show that users' phone call  patterns in $\SetSa$ can be matched to those in $\SetSb$ with a reasonably high accuracy.
%In the following we refer to this dataset by the CDR dataset.
\subsubsection{Dataset description and preprocessing}\label{sec: datasetdescription1}
\begin{figure}
\centering
\includegraphics[width=.45\columnwidth]{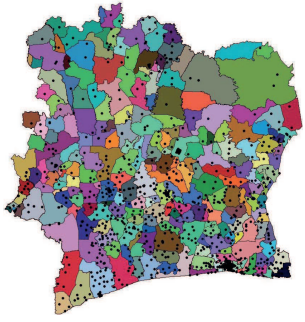}
\label{fig: dataset1}
\caption{(a) Position of Orange's GSM antennas in Ivory Coast~\cite{blondel2012data}. The sub-prefectures are shown by different colors.}
\end{figure}

The call-data records (CDR) dataset  consists of anonymized records of phone calls  between $50,000$ Orange customers (i.e., users)  in Ivory Coast~\cite{blondel2012data}, chosen randomly from millions of users.
%The CDR dataset consists of the call data records of  in the two-week period from  Monday $9^{\mbox{th}}$ to  Sunday $22^{\mbox{nd}}$ of April $2012$.
%The records were collected over the two-week period from  Monday $9^{\mbox{th}}$ to  Sunday $22^{\mbox{nd}}$ of April $2012$.
The dataset covers the two-week period from  Monday $9^{\mbox{th}}$ to  Sunday $22^{\mbox{nd}}$ of April $2012$ and contains the time of every call made by every user and the identifier of the antenna  to which the user was connected when making the call.
Figure~\ref{fig: dataset1} shows a map of Ivory Coast  with the positions of $1237$ antennas in the country indicated by  black circles~\cite{blondel2012data}.

%
%
%\begin{figure}
%\begin{minipage}[c]{0.35\columnwidth}
%\centering
%\vspace{0pt}
%\includegraphics[width=1\columnwidth]{FIG/ivory1.eps}
%\label{fig: dataset1}
%\end{minipage}
%%\end{figure}
%\hfill
%\begin{minipage}[c]{0.55\columnwidth}
%%\begin{table}
%\centering
%\vspace{0pt}
%\ra{1.2}
%\setlength{\belowrulesep}{0pt}
%\setlength{\aboverulesep}{0pt}
%\begin{tabular}{| >{\centering\arraybackslash}m{.9cm} | >{\centering\arraybackslash}m{.7cm} | >{\centering\arraybackslash}m{.7cm} | >{\centering\arraybackslash}m{.7cm} | }
%\toprule[1pt]
%Metric  &      \multicolumn{3}{c|}{Dataset} \tabularnewline
%\cmidrule{2-4}
% & CDR & WBH & GL \tabularnewline
% \midrule[0.5pt]
%  prop. &  $21.1\%$ & $90.1\%$ & $21.1\%$ \tabularnewline
%$l_1$    &  $18.6\%$ & $83\%$ & $21.1\%$ \tabularnewline
% cos &  $16.4\%$ & $77.5\%$ & $21.1\%$ \tabularnewline
%  dot &  $13.3\%$ & $70\%$ & $21.1\%$ \tabularnewline
%\bottomrule[1pt]
%\end{tabular}
%\caption{ }
%\label{table:res}
%%\end{table}
%\end{minipage}
%\end{figure}

We first split the CDR dataset  into two parts, where part one corresponds to the calls made in the first one-week period from the $9^{\mbox{th}}$ to $15^{\mbox{th}}$ of April, and part two corresponds to the calls made in the second one-week period from the $16^{\mbox{th}}$ to $22^{\mbox{nd}}$ of April.
%We refer to the  period of $9^{\mbox{th}}$ to $15^{\mbox{th}}$ of March by the \emph{first week} and  to the  period of $16^{\mbox{th}}$ to $22^{\mbox{nd}}$ of March by the \emph{second week}.
We then restrict our attention  only to users who are \emph{active} in both weeks, i.e., the users who made at least one call in each of the two weeks.
There are $\pop=46986$ such users, and overall they connected to $\staten=1211$  antennas.
Each user, on average, made  $101.2$ calls  and connected to $6.7$ different antennas.
We consider the coverage region of each antenna to be a location.
We disregard the timing information of the calls and construct the histograms of the calling patterns of each user in each week.
Thus, the histogram $\Gamma_{x_{\permU(i)}}$ (respectively, $\Gamma_{y_i}$) of user $i$ in the first (respectively, second) week gives the relative fractions of calls made by the user in various locations in the first (respectively, second) week.
The set  $\SetSa$ (respectively, $\SetSb$) consists of the histograms computed over the first week (respectively, second week).

\subsubsection{Matching accuracy with different metrics}\label{sec: expdeanony1}
%For every user $i$, $1 \leq i \leq \pop=46986$, the histogram  $\Gamma_{x_i}$ (respectively, $\Gamma_{y_i}$) is equal to the proportion of calls that user $i$ makes from GSM tower $l$, $1 \leq l \leq \staten=1211$, in the first week (respectively, second week).
After computing the histograms, we construct the complete bipartite graph $G$ shown in Figure~\ref{fig: bipartitematch4traj1} and described in Section~\ref{sec: proposedapproach1}.
%Graph $G$ for the CDR dataset  is shown in Figure~\ref{fig: bipartitematch4exp1}.
We choose edge weights $w_{\z i}$ given in~\eqref{eqn:weights} and  compute, by using (A1), a minimum-weight maximal matching on $G$.
The obtained result is shown in the first row of Table~\ref{table:resDatsets}.
Of $46986$ users, $9927$ are correctly matched, which gives an accuracy of $21.1\%$.
%We observe that although the underlying i.i.d. assumption of users' positions at every second of their trajectories is inherently false, the obtained accuracy is high considering the large number of users.
This means that, given the proportions of calls of users from different antennas during two consecutive weeks, we are able to correctly match more than one-fifth of them.
%Or from a privacy perspective, given anonymized temporal averages of all these users on one Monday, the identities of  close to  half of these users can be identified by tracking these users on a different Monday.
%\begin{figure}
%\centering
%\psfrag{x}[Bc][Bc]{\small{$\Gamma_{x_1}$}}
%\psfrag{b}[Bc][Bc]{\small{$\Gamma_{x_\pop}$}}
%\psfrag{y}[Bc][Bc]{\small{$\Gamma_{y_1}$}}
%\psfrag{d}[Bc][Bc]{\small{$\Gamma_{y_\pop}$}}
%\psfrag{w}[Bc][Bc]{} %{\small{$w_{11}$}}
%\psfrag{S}[Bc][Bc]{\parbox[c]{4.5cm}{\begin{center} $\SetSa$: Histograms of first \\  week ($9$ - $15$ April)\end{center}}}
%\psfrag{U}[Bc][Bc]{\parbox[c]{4.5cm}{\begin{center} $\SetSb$: Histograms of  second \\  week ($16$ - $22$ April)\end{center}}}
%\includegraphics[width=.5\columnwidth]{FIG/CDRbipartiteMatching.eps}
%\caption{We consider the scenario where anonymized call histograms of the users is released and an adversary who has  access to auxiliary histograms about the users tries to match the histograms that belong the same underlying user. The adversary builds the complete bipartite graph $G$ as described in Section~\ref{sec: proposedapproach1} and computes the matching corresponding the to the minimum-weight bipartite matching of the graph.}
% \label{fig: bipartitematch4exp1}
%\end{figure}
\begin{table}
\centering
\ra{1.2}
\setlength{\belowrulesep}{0pt}
\setlength{\aboverulesep}{0pt}
\begin{tabular}{| >{\centering\arraybackslash}m{.9cm} | >{\centering\arraybackslash}m{.7cm} |   >{\centering\arraybackslash}m{.7cm} | >{\centering\arraybackslash}m{.9cm} | >{\centering\arraybackslash}m{.85cm} | >{\centering\arraybackslash}m{.85cm} | >{\centering\arraybackslash}m{.9cm} |}
\toprule[1pt]
Dataset    &    \multicolumn{2}{c|}{Characteristics} &    \multicolumn{4}{c|}{Choice of metric in (A1)} \tabularnewline
\cmidrule{2-7}
 & $\pop$ & $\staten$ & proposed & $l_1$ & cosine & dot  \tabularnewline
\midrule[0.5pt]
CDR & $46986$ & $1211$ & $21.1\%$ & $18.6\%$ & $16.4\%$  & $13.3\%$  \tabularnewline
\midrule[0.25pt]
WBH & $121$ & $83219$ & $90.0\%$ & $81.1\%$ & $72.7\%$  & $64.4\%$  \tabularnewline
\midrule[0.25pt]
GL & $154$ & $1024$ & $58.4\%$ & $51.3\%$ & $52.0\%$  & $46.8\%$  \tabularnewline
\bottomrule[1pt]
\end{tabular}
\vspace{3pt}
\caption{Matching accuracy obtained on $G$ in Figure~\ref{fig: bipartitematch4traj1} by using (A1) with  various choices for the distance/similarity measures between the histograms defined in ~\eqref{eqn:weights},~\eqref{eq: Ch4lpdist1},~\eqref{eq: Ch4dotsim1}, and~\eqref{eq: Ch4cossim1}. The proposed weight function consistently yields  the highest  accuracy for all three datasets.}
\label{table:resDatsets}
\vspace{-5pt}
\end{table}
We also compare the matching accuracy obtained by using the distance measure~\eqref{eqn:weights} with the accuracy obtained by using the distance measures  given in~\eqref{eq: Ch4cossim1} and~\eqref{eq: Ch4lpdist1}, as well as the similarity measure of~\eqref{eq: Ch4dotsim1}.
%We remark that the dot product given in~\eqref{eq: Ch4dotsim1} is  actually a similarity measure.
%In these alternative approaches, the matching chosen is the permutation defined by the minimum-weight maximal bipartite matching for the case of distance measures and by the maximum-weight bipartite matching for the case of the similarity measure.
%The obtained accuracy for each choice of weight function is shown in Figure~\ref{fig: experothermeasu1}.
We observe from the table that the matching accuracy obtained by using the weight function proposed in~\eqref{eqn:weights} is significantly higher than that obtained by using any of the other \emph{heuristic} measures.
We remark that the naive approach of deciding on a purely random matching between the histograms yields, on average, one correctly matched user.
%The , and thus an accuracy of $0.002\%$.
The resulting accuracy ($0.002\%$) is negligible compared to those obtained in Table~\ref{table:resDatsets}.

\subsubsection{Effect of varying the number $\pop$ of users}\label{sec: expervaryingN1}
\begin{figure}
\centering
\psfrag{A}[Bc][Bc]{\small{Accuracy ($\%$)}}
\psfrag{x}[Bc][Bc]{\small{$\pop$}}
\psfrag{S}[Bc][Bc]{\small{$\SetSa$}}
\psfrag{U}[Bc][Bc]{\small{$\SetSb$}}
\psfrag{W}[Bc][Bc]{\small{$|\SetSa|=|\SetSb|=\pop$}}
\psfrag{p}[l][l]{\footnotesize{\text{proposed}}}
\psfrag{l}[l][l]{\footnotesize{\text{$l_1$-norm}}}
\psfrag{c}[l][l]{\footnotesize{\text{cosine}}}
\psfrag{d}[l][l]{\footnotesize{\text{dot}}}
\includegraphics[width=1\columnwidth]{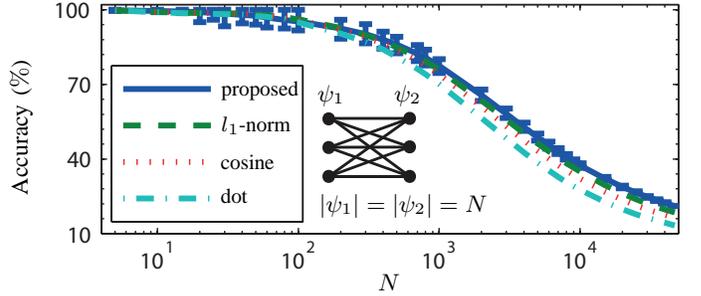}
\caption{The obtained average accuracy by using different edge weights as a function of the number of users $\pop$, in the setting where  the histograms in sets $\SetSa$ and $\SetSb$  are generated by the same set of $\pop$ users (i.e., $\USetSa=\USetSb$). 
The measures are defined in~\eqref{eqn:weights},~\eqref{eq: Ch4lpdist1},~\eqref{eq: Ch4cossim1}, and~\eqref{eq: Ch4dotsim1}.
The  $90\%$ confidence interval is also shown for our proposed metric.
  We observe that  increasing the number of users $\pop$ leads to  a reduction in the matching accuracy.}
\label{fig: expervarN1}
\end{figure}

%In the previous experiment, the number $\pop$  of users in graph $G$ in Figure~\ref{fig: bipartitematch4traj1} was $46986$.
In this experiment,
%we study the effect of varying this number on the matching accuracy. %de-anonymize all the
%In other words,
we keep $\USetSa=\USetSb$ but vary $|\USetSa|$.
%We remark that in this experiment, $|\SetSa|=|\SetSb|$ and  the labeled and the unlabeled histograms are generated by the same set users.
%We consider the same dataset as in the experiment in Section~\ref{sec: expdeanony1}.
We first choose uniformly at  random a subset of the  $46986$ users considered in the previous experiment.
We denote the subset size by $\pop$.
We then choose sets $\SetSa$ and $\SetSb$ to be  the histograms associated with   the $\pop$ chosen users in the first week and the  second week, respectively.
%After computing  $\Gamma_{x_i}$ and $\Gamma_{y_i}$ for every chosen user,
%We construct the complete bipartite graph  $G$ with edge weights given in~\eqref{eqn:weights} and use (A1).
We then apply (A1) to the graph  $G$ of Figure~\ref{fig: bipartitematch4traj1} with  different choices of edge weights. %edge weights given in~\eqref{eqn:weights}.
%, and  compute a minimum-weight matching on the graph.
%We then match the chosen users in the sets $S_1$ and $S_2$ by constructing the complete bipartite graph with edge weights given in~\ref{} and performing a minimum weight matching.
%We vary the size of the chosen random subset $\pop$ from $1000$ to $46986$, where for each value of $\pop$ we repeat the experiment several times.
For each value of $\pop$ we repeat the experiment several times, choosing  the subset randomly and performing the matching.
The obtained average accuracies  are shown in Figure~\ref{fig: expervarN1} as a function of $\pop$ for each choice of edge weight.
%We remark that in this experiment, $|\SetSa|=|\SetSa|=\pop$ and the histograms are generated by the same set of $\pop$ users.
%  for each value of $\pop$,  the sets $\SetSa$ and $\SetSb$  contain the trajectories of the same $\pop$ users; in other words, the histograms in sets $\SetSa$ and $\SetSb$  are generated by the same set of $\pop$ users.
We observe from Figure~\ref{fig: expervarN1} that as the value of $\pop$  increases, the matching accuracy under all  metrics decreases. %of users in the both sets $\SetSa$ and $\SetSb$
This is expected because as $\pop$ increases,  the habits of the users start resembling those of others, and  it becomes more difficult to distinguish the histograms of one user from those of others.
%histograms become closer to each other and  it becomes more difficult to distinguish the histograms from each other.
Hence,  the matching accuracy decreases.
Furthermore, although the $21.1\%$  accuracy obtained with the proposed metric of~\eqref{eqn:weights} in Table~\ref{table:resDatsets} might seem small at first, it is associated with a large value of $\pop$.
If the number  of users were smaller,  the  accuracy would be higher (e.g., $78\%$ for $1000$ users).

\subsubsection{Matching different subsets of users}\label{sec: expsingleton1}
Following the discussion in Section~\ref{sec: uncbipartitematch1}, here we investigate the practical scenario where the histograms in sets $\SetSa$ and $\SetSb$ belong to different sets of distinct users.
In other words, in this experiment $\USetSa\neq \USetSb$.
We first consider the setting in which we are given  histograms of all users on the second  week  but only a subset of  users on the  first week.
That is, $\USetSa \subset \USetSb$, as depicted  in Figure~\ref{fig: bipartitematchsubset}.
%The objective is to match the users' unlabeled histograms from the first week to the correct histograms from the second week.
%This is the setting depicted  in Figure~\ref{fig: bipartitematchsubset}
%We proceed as follows.

\begin{figure}
\centering
\psfrag{A}[Bc][Bc]{\small {Accuracy ($\%$)}}  %{$\#$ corr. matches $/$ $|\SetSa|$ ($\%$)}}
\psfrag{B}[Bc][Bc]{\small{$\#$ corr. matches}}
\psfrag{x}[Bc][Bc]{\small{Number of histograms in the first dataset ($|\SetSa|$)}}
\psfrag{S}[Bc][Bc]{\small{$\SetSa$}}
\psfrag{U}[Bc][Bc]{\small{$\SetSb$}}
\psfrag{L}[l][l]{\scriptsize{Accuracy}}
\psfrag{G}[l][l]{\scriptsize{$\#$ corr. matches}}
%\psfrag{W}[Bc][Bc]{\small{$|\SetSa|=|\SetSb|=\pop$}}
\psfrag{W}[Bc][Bc]{\parbox[c]{2cm}{\small{ $|\SetSb|=\pop$ \\ $|\SetSa|<|\SetSb|$}}}
\includegraphics[width=1\columnwidth]{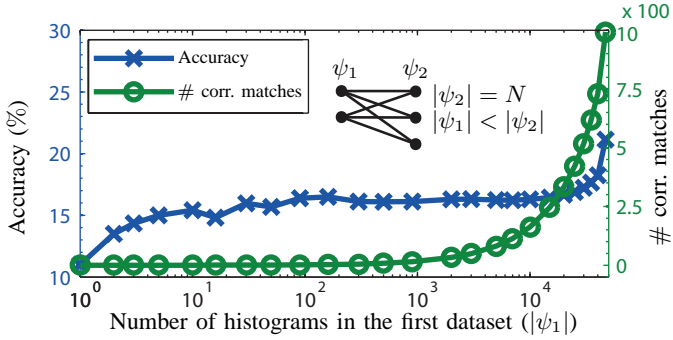}
\caption{The average number of correct matches and the  average accuracy  as a function of $|\SetSa|$ in the setting where we are given  histograms of all users on the second week (i.e., $|\SetSb|=\pop$) but only a subset of  users on the  first week (i.e., $|\SetSa|<\pop$).
The leftmost point represents one-by-one user matching approach, which yields the smallest accuracy.}
\label{fig: expervarN2} % The results are averaged over several repetitions of the experiment.
\end{figure}

We let $\SetSb$ be the collection of histograms of all the $\pop=46986$ users on the second week.
For $\SetSa$ we use the collection of histograms of a randomly chosen subset of users on the first week.
%We construct the complete bipartite graph $G$ illustrated in Figure~\ref{fig: bipartitematchsubset}.
%We then apply (A1) to the graph  $G$ of Figure~\ref{fig: bipartitematch4traj1} with edge weights given in~\eqref{eqn:weights}.
We construct $G$  in Figure~\ref{fig: bipartitematchsubset} with edge weights in~\eqref{eqn:weights} and  run (A1).
%as described in Section~\ref{sec: proposedapproach1} (refer to Figure~\ref{fig: bipartitematch4exp1}), where now the left part of the graph  has $|\SetSa|=\pop$ nodes whereas the right part of the graph  has $|\SetSb|<\pop$ nodes.
%We choose edge weights given in~\eqref{eqn:weights} and  use (A1). % compute a minimum-weight maximal matching on the graph.
The resulting matching has a size equal to $|\SetSa|$.
The number of correctly matched histograms  in the set $\SetSa$ divided by $|\SetSa|$ defines the obtained accuracy.
%We repeated this experiment several times where in each iteration we chose randomly the subset of histograms in the second week.
%The results are averaged over several repetitions of the experiment.
Figure~\ref{fig: expervarN2} shows the average number of correct matches and  the average  accuracy obtained for different values of $|\SetSa|$, where the results are averaged over several repetitions of the experiment.
%[RRRR: zero deviation for the two extreme points].
%The extreme case when $|\SetSb|=1$ corresponds to the scenario where the attacker is interested in matching only one user's histogram of the second week to his histogram of the first week.
%Let $\Gamma_{y_i}$ denote the histogram of the user in the second week.
%For this case, the right part of graph $G$ (refer to Figure~\ref{fig: bipartitematchsubset}) consists of only one node (i.e., $\SetSb=\{\Gamma_{y_i}\}$).
%The solution to the minimum-weight bipartite matching is the histogram  $\Gamma_{x_z}$ from the first week that gives the minimum value of the weight $w_{zi}$ computed in~\eqref{eqn:weights}.
%%We computed the solution where if  there existed ties (i.e., multiple users in $\SetSa$ having minimum weight), we break them randomly.
%We estimate the average probability of correct matching under this procedure by computing the fraction of choices of $\SetSb$ that lead to correct matchings.
%If there exist ties (i.e., multiple users in $\SetSa$ having minimum weight), we break them randomly.
%The obtained average of correct matches is $9.5\%$, which is much smaller than the matching accuracy of $21.1\%$ in Figure~\ref{fig: experothermeasu1}.
%While the size of the set $\SetSa$ of unlabeled histograms remains constant, as the set $\SetSb$ of labeled trajectories  becomes larger,
The leftmost point represents one-by-one user matching approach, which yields the smallest accuracy.
From a user's perspective, as $|\SetSa|$ increases,  the adversary has more information available and thus can obtain a better matching.
%As $|\SetSa|$ increases,  we have more information available and thus can do a better matching.
Hence, the obtained matching accuracy increases.
 % as we traverse to the right, the set $\USetSa \cap \USetSb$ increases (i.e., while the total number of users $\USetSa \cup \USetSb$ remains% fixed.
%This is expected because in the latter case we have more data (the set of labeled trajectories $\SetSb$ is larger) and thus we can match better.
This observation has important implications in the perspective of privacy of anonymized  statistics.
A user's privacy depends not only on how much her trajectory is revealed to the adversary, but also on how much of others' trajectories are revealed to the adversary.

In the second part of this experiment, we consider the setting where  $|\USetSa \cap \USetSb|=r < \pop$.
This is the setting depicted  in Figure~\ref{fig: bipartitematchunc1}.
We choose uniformly at random a set of histograms from the first week and from the second week, such that  $|\USetSa|=|\USetSb|=5000$, and $|\USetSa \cap \USetSb|=3750$.
We choose these values as an example.
We then construct $G$ in Figure~\ref{fig: bipartitematchunc1} with edge weights given in~\eqref{eqn:weights}.
We first choose $3750$ of the unlabeled histograms in $\USetSa$ and matched them to $3750$  of the labeled histograms in $\USetSb$, such that the summation of the distance between the matched pairs  is minimized.
We do this by applying (A2) with $r=3750$ to $G$.
%After running the algorithm, $1340$ out of the $3750$ chosen unlabeled histogram are correctly matched and $3750-1340=2410$ of them are incorrectly matched.
%Hence, in total  $3750-1340=2410$ of the unlabeled histograms are incorrectly matched.
Alternatively, we  match all the $5000$ unlabeled histograms in $\USetSa$ to   the labeled histograms in $\USetSb$ by applying (A1) to $G$.
%Then, $1672$ out of the $5000$ unlabeled histograms are correctly matched and $5000-1672=3328$ of them are incorrectly matched.
The obtained results are shown in Table~\ref{tab: singletonres1}.
Although the first approach yields a smaller number of correct matches ($1340$ versus $1672$) compared to the second approach,  it yields a larger percentage of correct matches ($36\%$ versus $33\%$).
Therefore, it makes sense to use (A2) instead of (A1)  when the adversary is interested in maximizing his percentage accuracy (i.e., number of correct matches divided by the size of the outputted matching). %number of incorrect matches.
\begin{table}%[h]
\centering
\ra{1.2}
\setlength{\belowrulesep}{0pt}
\setlength{\aboverulesep}{0pt}
\begin{tabular}{| >{\centering\arraybackslash}m{1.5cm} | >{\centering\arraybackslash}m{1.2cm} | >{\centering\arraybackslash}m{1.2cm} |>{\centering\arraybackslash}m{1.2cm} | >{\centering\arraybackslash}m{1.4cm} | }
\toprule[1pt]
Algorithm    &   matching size & $\#$ correct matches & Percentage accuracy & $\#$ incorrect matches \tabularnewline
\midrule[0.5pt]
(A2) with $r=3750$ & 3750 & $1340$ & $36\%$ & $2410$ \tabularnewline
\midrule[0.25pt]
(A1) & 5000 & $1672$ & $33\%$ & $3328$ \tabularnewline
\bottomrule[1pt]
\end{tabular}
\vspace{3pt}
\caption{Obtained matching result for the case $|\USetSa \cap \USetSb|=r < \pop$ depicted  in Figure~\ref{fig: bipartitematchunc1} with $r=3750$ and $\pop=5000$. Compared to the second approach, the first approach yields a smaller number of correct matches but a larger percentage of correct matches.}
\label{tab: singletonres1}
\vspace{-5pt}
\end{table}

\subsubsection{Effect of varying the time-duration of data collection}\label{sec: expervaryingdays1}
\begin{figure}
\centering
\psfrag{A}[Bc][Bc]{\small{Accuracy ($\%$)}}
\psfrag{M}[Bc][Bc]{\small{Time duration included in the dataset}}
\psfrag{a}[Bc][Bc]{\footnotesize{Mon}}
\psfrag{b}[Bc][Bc]{\footnotesize{Mon--Tue}}
\psfrag{c}[Bc][Bc]{\footnotesize{Mon--Wed}}
\psfrag{d}[Bc][Bc]{\footnotesize{Mon--Thu}}
\psfrag{e}[Bc][Bc]{\footnotesize{Mon--Fri}}
\psfrag{f}[Bc][Bc]{\footnotesize{Mon--Sat}}
\psfrag{g}[Bc][Bc]{\footnotesize{Mon--Sun}}
\psfrag{S}[Bc][Bc]{\small{$\SetSa$}}
\psfrag{U}[Bc][Bc]{\footnotesize{$\SetSb$}}
\psfrag{W}[Bc][Bc]{\footnotesize{$|\SetSa|=|\SetSb|=\pop$}}
\psfrag{p}[l][l]{\footnotesize{\text{proposed}}}
\psfrag{l}[l][l]{\footnotesize{\text{$l_1$-norm}}}
\psfrag{C}[l][l]{\footnotesize{\text{cosine}}}
\psfrag{D}[l][l]{\footnotesize{\text{dot}}}
\includegraphics[width=1\columnwidth]{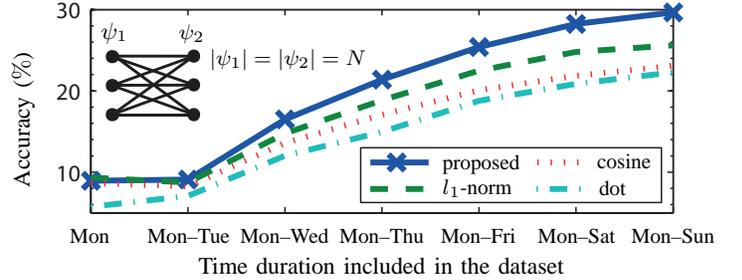}
\caption{The obtained matching accuracy ($\pop=30937$) with different choices of edge weights as a function of time-duration over which users' statistics are computed.
The measures are defined in~\eqref{eqn:weights},~\eqref{eq: Ch4lpdist1},~\eqref{eq: Ch4cossim1}, and~\eqref{eq: Ch4dotsim1}.  As long as users' habits remain stationary and ergodic, by increasing the time-duration over which statistics are computed, histograms belonging to each user become closer to each other, and thus the overall matching accuracy increases.}
\vspace{-5pt}
\label{fig: experdayvar1}
\end{figure}

We now investigate how the matching accuracy is affected by the time-duration  over which users' statistics are computed.
%can be improved by using statistics of users from multiple days of the week (i.e., by using histograms belonging to longer time-periods).
We  consider all users who  were  active on each Monday of the two-week period, i.e., users who made at least one call on  Monday $9^\text{th}$ and  on Monday $16^\text{th}$ of April.
There are $\pop=30937$ such users.
In the first part of this experiment, the set  $\SetSa$ (respectively, $\SetSb$) corresponds to the histograms of the number of calls of this $\pop$ users from the $\staten$ locations (i.e., GSM antennas) during the first Monday (respectively, second Monday).
We then  construct graph  $G$  illustrated in Figure~\ref{fig: bipartitematch4traj1} with  different choices of edge weights, and  run (A1).
The obtained accuracy, marked on the x-axis by ``Mon'',  is shown in Figure~\ref{fig: experdayvar1}.

In the second part of this experiment, we  increase the time-duration over which we compute users' statistics.
We compute the statistics of the same $\pop$ users during the  Monday and Tuesday of the first week and  of the second week.
Thus, the set  $\SetSa$ (respectively, $\SetSb$) now corresponds to the histograms of the number of calls of the $\pop$ users from the $\staten$ locations during the first (respectively, second) Monday and Tuesday.
We then construct the graph $G$ with  different choices of edge weights and run (A1).
The obtained matching accuracy, marked by   ``Mon-Tue'',  is shown in Figure~\ref{fig: experdayvar1}.
Similarly, we increase the number of considered days for every user and repeat the experiment.
These results are shown in the figure as well.
As can be seen from Figure~\ref{fig: experdayvar1}, the matching accuracy increases as we include more days in the dataset.
%This is because we have more information available for matching the users, specifically multiple days' call statistics instead of two days.
This is because as long as users' habits remain stationary and ergodic, by increasing the time-duration over which statistics are computed, the two histograms belonging to each user become closer to each other, and thus the overall matching accuracy increases.
Furthermore,    the matching accuracy obtained by using the weight function proposed in~\eqref{eqn:weights} is significantly higher than that obtained by using any of the other \emph{heuristic} measures.
A standout feature in Figure~\ref{fig: experdayvar1} is the fact that the incremental improvement in going from Mon to Mon-Tue is lower than that observed in other data points in the graph.
This is probably because  Monday, April $9^\text{th}$ was \emph{White Monday}, a public holiday in Ivory Coast.
On the following day (i.e., Tuesday  $10^\text{th}$ of April) the users made on average only $1.9$ calls compared to the average of  $7.2$ calls per day.
%Consequently, the matching accuracy has only a slight improvement when we include Tuesday's statistics for the matching.

\subsubsection{Effect of location aggregation}\label{sec: experlocobfus1}
%For some applications, it is not necessary to have  high spatial resolution (i.e., fine location granularity) in the mobility statistics.
%[RRRR: For example, having  coarse  spatial granularity (e.g., $50$ m) compared to  fine spatial granularity (e.g., $10$ m obtained via GPS)  is sufficient for an urban planning application where the objective is to find the most popular location for building a department store. ]
In addition to the removal of user identifiers (i.e., anonymization), an additional  well-known  privacy-protection mechanism that is usually applied to mobility traces is spatial-resolution reduction, which is known also as location obfuscation or location aggregation~\cite{huang2010preserving,christin2011survey}.
Here we investigate the effect of location aggregation on the matching accuracy.
%Here we apply location aggregation to the CDR dataset  and we compute the matching accuracy on the corresponding histograms.

The Orange call-data records dataset also includes a low-spatial resolution version~\cite{blondel2012data} that contains the time of every call made by $500,000$ randomly chosen users and the sub-prefectures (i.e., administrative divisions within the provinces) of the antennas to which they were connected while making the call.
%where the call records  of $500,000$ randomly chosen users is provided, and where the spatial resolution is reduced by publishing the sub-prefectures (i.e., administrative divisions within the provinces) of antennas rather than the antennas' identifiers.
The sub-prefectures, shown by different colors in Figure~\ref{fig: dataset1}, in general contain multiple antennas, thus the dataset has a  spatial resolution lower than the original dataset.
We consider a two-week period  and randomly choose a subset of size $\pop=46986$ active users out of the total $500,000$ users.
The set  $\SetSa$ (respectively, $\SetSb$) corresponds to the histograms of the number of calls of the $\pop$ users from each sub-prefectures (i.e., location) during the first week (respectively, second week).
Users, in total, made calls from $\staten=237$ sub-prefectures. %, which has a much smaller value than the $1211$ antennas.
We then  construct the complete bipartite graph  $G$  illustrated in Figure~\ref{fig: bipartitematch4traj1} with edge weights given in~\eqref{eqn:weights}, and  run (A1).
There are $2070$  correctly matched users, which gives an accuracy of $4.40\%$. %\notes{Did you do any averaging over random realizations?}
The obtained accuracy is much lower than the $21.1\%$ obtained  for the same number of users in the original high-resolution dataset.
As antennas are aggregated into sub-prefectures, users' histograms become less distinguishable and, as a result, the matching accuracy drops significantly.

\subsection{Experiments on web browsing history (WBH) dataset}\label{sec: webbrowsinghistorydataset}
%%In section~\ref{sec: CDRdataset}, we performed experiments on a mobility-related dataset.
%In this subsection we perform experiments on the Web History Repository~\cite{herder2011experiences}, which   consists of anonymized detailed web browsing history of hundreds of users.
%Users can upload their anonymized usage data to the repository by using a Mozilla Firefox add-on.
%%This dataset  consists of anonymized detailed records of phone calls  between Orange customers (i.e., users)  in Ivory Coast.
%In order to protect the users' privacy, all urls and hosts are represented by a global unique identifier.
%%In our experiments in this section, the sets  $\SetSa$ and $\SetSb$ correspond to the histograms of the number of visits of users to different websites (i.e., locations) measured over two different non-overlapping time periods.
%%Using \emph{only} the histogram of visits  of the various users to  various locations in each time-period, we show that users' web browsing patterns in $\SetSa$ can be matched to those in $\SetSb$ with a high accuracy.
%In the following we refer to this dataset by the WBH dataset.
\subsubsection{Dataset description and preprocessing}\label{sec: WBHdescription1}
The Web History Repository~\cite{herder2011experiences} consists of anonymized detailed web browsing history of hundreds of users.
Users can upload their anonymized usage data to the repository by using a Mozilla Firefox add-on.
%This dataset  consists of anonymized detailed records of phone calls  between Orange customers (i.e., users)  in Ivory Coast.
In order to protect the users' privacy, all URLs and hosts are represented by a global unique identifier.
%In our experiments in this section, the sets  $\SetSa$ and $\SetSb$ correspond to the histograms of the number of visits of users to different websites (i.e., locations) measured over two different non-overlapping time periods.
%Using \emph{only} the histogram of visits  of the various users to  various locations in each time-period, we show that users' web browsing patterns in $\SetSa$ can be matched to those in $\SetSb$ with a high accuracy.
%We refer to this browsing history dataset by the WBH dataset.
The web browsing history (WBH) dataset contains the browsing history of $472$ users.
Users participated in the data collection for different time-periods during the course of several years.
For each user, the dataset contains every visited URL (with encrypted name), the \emph{favicon} identifier associated with the URL, and the time of visit to the URL.
The favicon, also known as a shortcut icon, is a small icon associated with a particular website.
%Generally,  different urls (e.g., ``facebook.com/1.html'' and ``facebook.com/2.html'') associated with the same website (e.g., ``facebook.com'') would have the same favicon and thus can be mapped to a single website.
Generally,  different URLs  associated with the same website (e.g., domain name)  have the same favicon and hence can be mapped to a single website.
For example, if a user visits the URLs ``news.yahoo.com'' and ``mail.yahoo.com'', the URLs will appear with different encrypted names in the database; however, both URLs will have the same favicon identifier (e.g., ``1'').
Thus, we can learn that the user has visited a particular website (i.e., ``yahoo.com'') twice.
%Hence, we treat favicon identifier as the website identifier.

%In the database, the favicon identifier of the urls that do not have a favicon is set to zero.

\begin{figure}
\centering
\psfrag{A}[Bc][Bc]{\small{Number of visits}}
\psfrag{N}[Bc][Bc]{\small{Website identifier}}
\includegraphics[width=1\columnwidth]{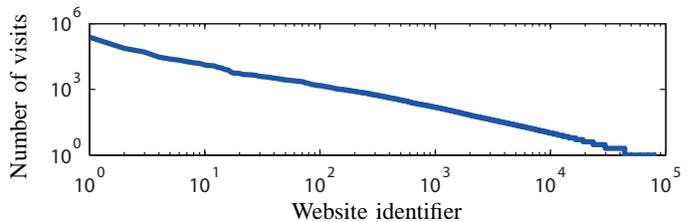}
\caption{The total number of visits (i.e., popularity) to the $\staten$ websites by all the users  in the two-week period. The figure is plotted in a log-log scale and the websites are indexed according to their popularity.}
\label{fig: WBHpopularlocs1}
\end{figure}

We remove from the dataset all URLs that do not have a favicon.
We consider each website (e.g., ``yahoo.com'') to be a location and treat the favicon identifier as the website identifier for each URL.
%a particular favicon represents a single website, such as facebook.com.
%Hence, a particular favicon represents a single website, such as facebook.com.
%In the database, the favicon identifier of the urls that do not have a favicon is set to zero.
We then  identify the period of  two consecutive weeks that has the maximum number of active users (i.e., users who visit at least one website during each of the two weeks).
There are $\pop=121$ active users in this two-week period.
%We then discarded all the visited urls by the users that have favicon identifier equal to zero.
They  visited $\staten=83219$ different websites, $77935$ of which were visited by not more than one user.
Figure~\ref{fig: WBHpopularlocs1} shows a log-log plot of the total number of visits to the websites by all the users in the two-week period.
The y-axis values represent the \emph{popularity} of the websites.

We disregard the timing information of the visited websites and construct the  histograms of the browsing patterns of each user in each week.
Thus, the histogram $\Gamma_{x_{\permU(i)}}$ (respectively, $\Gamma_{y_i}$) of user $i$ in the first (respectively, second) week gives the relative fractions of the visits to various websites by that user in the first (respectively, second) week.
The set  $\SetSa$ (respectively, $\SetSb$) consists of the histograms computed over the first week (respectively, second week).
%The set  $\SetSa$ (respectively, $\SetSb$) corresponds to the histograms of the number of visits of the $\pop$ users to the $\staten$ websites  during the first week (respectively, second week).
%We consider the scenario where call histograms in the form of $\SetSa$  have been released; and an attacker who has  access to auxiliary histograms in the form of $\SetSb$ tries to match the histograms that belong the same underlying user.
%Given $\SetSa$ and $\SetSb$, the goal of the matching task is to match every histogram in $\SetSb$ to the histogram in $\SetSa$ produced by the same user, or equivalently, to estimate the permutation $\permU$ in~\eqref{eq:Ch4perturb1}.

\subsubsection{Matching accuracy with different metrics}\label{sec: WBHmatchingexp1}
%We then construct the complete bipartite graph $G$ shown in Figure~\ref{fig: bipartitematch4traj1}, and apply the matching algorithms proposed in Sections~\ref{sec: Ch4problemstatement} and~\ref{sec: uncbipartitematch1} on this graph with appropriately chosen edge-weights.
%After computing the histograms,
We construct the graph $G$ shown in Figure~\ref{fig: bipartitematch4traj1} from the histograms and apply (A1) to $G$ with different choices of edge weights.
The obtained results are shown in the second row of Table~\ref{table:resDatsets}.
% by using the proposed metric in~\eqref{eqn:weights}, $l_1$-norm in~\eqref{eq: Ch4cossim1}, cosine distance in~\eqref{eq: Ch4lpdist1}, and  the dot product in~\eqref{eq: Ch4dotsim1} was $90\%$, $83\%$, $77.5\%$, and $70\%$, respectively.
We observe that   the matching accuracy obtained by using the weight function proposed in~\eqref{eqn:weights} is significantly higher than that obtained by using any of the other \emph{heuristic} measures.
Furthermore, given the proportions of visited websites    during two consecutive weeks, we are able to correctly match almost all of them.

\subsubsection{Considering  popular websites}\label{sec: WBHpopularexp1}
\begin{figure}
\centering
\psfrag{p}[l][l]{\footnotesize{\text{proposed}}}
\psfrag{l}[l][l]{\footnotesize{\text{$l_1$-norm}}}
\psfrag{c}[l][l]{\footnotesize{\text{cosine}}}
\psfrag{d}[l][l]{\footnotesize{\text{dot}}}
\psfrag{A}[Bc][Bc]{\small{Accuracy ($\%$)}}
\psfrag{N}[Bc][Bc]{\small{Number of websites ($\staten$)}}
\includegraphics[width=1\columnwidth]{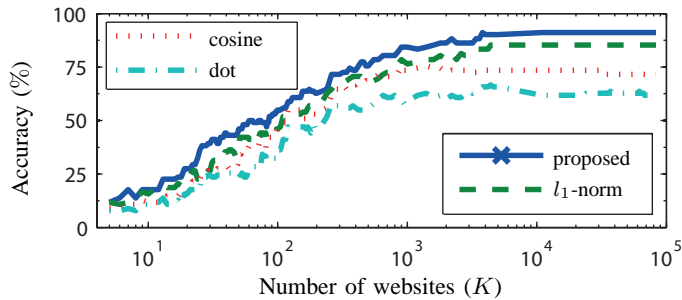}
\caption{Matching accuracy for the WBH dataset with $\pop=102$ users by using different measures when only a subset of the popular websites are considered.
The measures are defined in~\eqref{eqn:weights},~\eqref{eq: Ch4lpdist1},~\eqref{eq: Ch4cossim1}, and~\eqref{eq: Ch4dotsim1}.
%The measures are as follows: proposed~\eqref{eqn:weights}, $l_1$-norm~\eqref{eq: Ch4lpdist1}, cosine distance~\eqref{eq: Ch4cossim1}, and dot product~\eqref{eq: Ch4dotsim1}.
The proposed weight function yields the highest percentage accuracy in the matching. The popularity of the websites is shown in Figure~\ref{fig: WBHpopularlocs1}. }
\label{fig: WBHpopularexp1}
\end{figure}

One reason we  obtain a high matching accuracy is that some websites are  visited by only a small number of users during the two-week period,  hence it is easy to match those users.
%One reason for obtaining a high matching accuracy is the fact that some websites have been visited by a small subset of users (e.g., only one user) during the two-week period (for example, due to the ethnicity of the visiting users), and hence it would be easy to match those subset of users.
%This stems from the very large number of websites that exist and that each user can visit.
%This is in contrast to the previous dataset where the total number of locations where much smaller (i
%For example, a user might visit a website that is local to his geographical location or ethnicity.
We investigate this effect as follows.
We  consider all users who  visited at least one of the top $5$ popular websites, in Figure~\ref{fig: WBHpopularlocs1}.
There are $\pop=102$ such users.
We consider a subset (of size not less than $5$) of the most popular of the visited websites (refer to Figure~\ref{fig: WBHpopularlocs1}).
We then keep for every user $i$ ($ 1 \leq i \leq 102$) the elements of $\Gamma_{x_{\permU(i)}}$ and $\Gamma_{y_i}$ that correspond to the considered subset of websites, and we set the remaining elements equal to zero.
%We discard users for whom the resulting histogram $\Gamma_{x_{\permU(i)}}$ or $\Gamma_{x_{\permU(i)}}$ is the all-zero vector.
We then re-normalize the remaining histograms such that they sum to one.
We reconstruct, by using different choices of edge weights, the  bipartite graph $G$ in Figure~\ref{fig: bipartitematch4traj1}  and run (A1)  on the graph.
We repeat the experiment by varying the size of the considered subset of popular websites.
The result is shown in Figure~\ref{fig: WBHpopularexp1}.
As expected, as fewer websites are considered, we have less information available for matching;  hence the matching accuracy drops.
However, by considering merely the top $60$ most popular websites, we can still correctly match more than $50\%$ of users.
Moreover, as in Table~\ref{table:resDatsets}, the matching accuracy obtained by using the weight function in~\eqref{eqn:weights} is consistently higher than that obtained by using any of the other \emph{heuristic} measures.
%[RRRR: number of users]

%%%%%%%%%%%%%%%%%%%%%%%%%%%%%%%%%%%%%%%%%%%%%%%%%%%%%%%%%%%%%%%%%
%%%%%%%%%%%%%%%%%%%%%%%%%%%%%%%%%%%%%%%%%%%%%%%%%%%%%%%%%%%%%%%%%
%%%%%%%%%%%%%%%%%%%%%%%%%%%%%%%%%%%%%%%%%%%%%%%%%%%%%%%%%%%%%%%%%
%%%%%%%%%%%%%%%%%%%%%%%%%%%%%%%%%%%%%%%%%%%%%%%%%%%%%%%%%%%%%%%%%
\subsection{Experiments  on GeoLife (GL) GPS dataset}\label{sec: GeoLifedataset}
%In this subsection we perform experiments on the Geolife dataset~\cite{zheng2008understanding}, which contains the GPS traces of $182$ users collected over five years.
%A GPS trajectory in this dataset is represented by a sequence of time-stamped points, each of which contains the information of latitude, longitude and altitude.
%The trajectories are wildly distributed over many cities in China and even some in  the USA and Europe, but the majority of the data is created in the city of Beijing.
%In the following we refer to this dataset by the GL dataset.
%[RRRR: statistics]
\subsubsection{Dataset description and preprocessing}\label{sec: geolifedescription1}
\begin{figure}
\centering
\subfigure[]{
\psfrag{L}[Bc][Bc]{\small{Latitude}}
\psfrag{G}[Bc][Bc]{\small{Longitude}}
\includegraphics[width=.46\columnwidth]{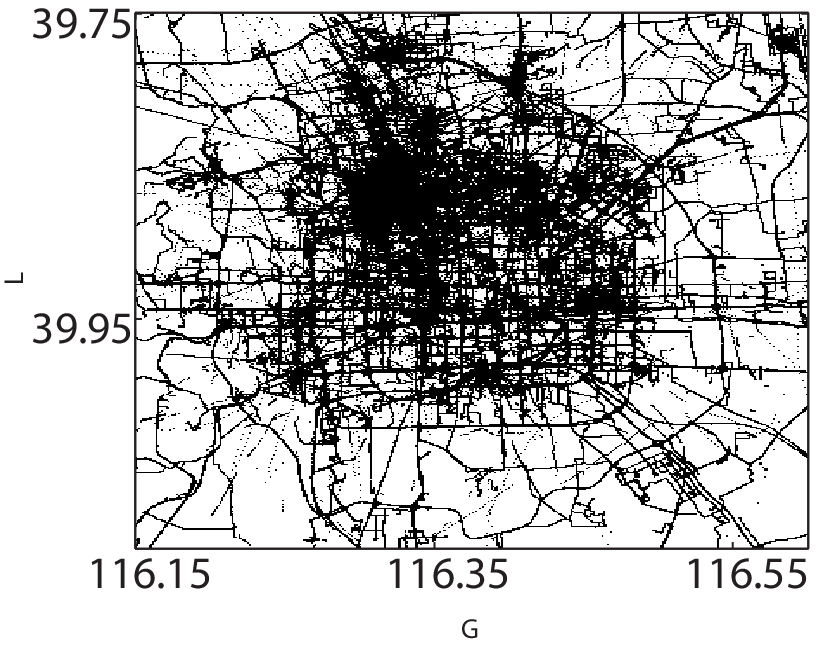}
\label{fig: geolifemap1}
}
\hfill
\subfigure[]{
\psfrag{U}[Bc][Bc]{\small{User ID}}
\psfrag{B}[Bc][Bc]{\footnotesize{2008}}
\psfrag{D}[Bc][Bc]{\footnotesize{2010}}
\psfrag{F}[Bc][Bc]{\footnotesize{2012}}
\psfrag{T}[Bc][Bc]{\small{Time}}
\includegraphics[width=.46\columnwidth]{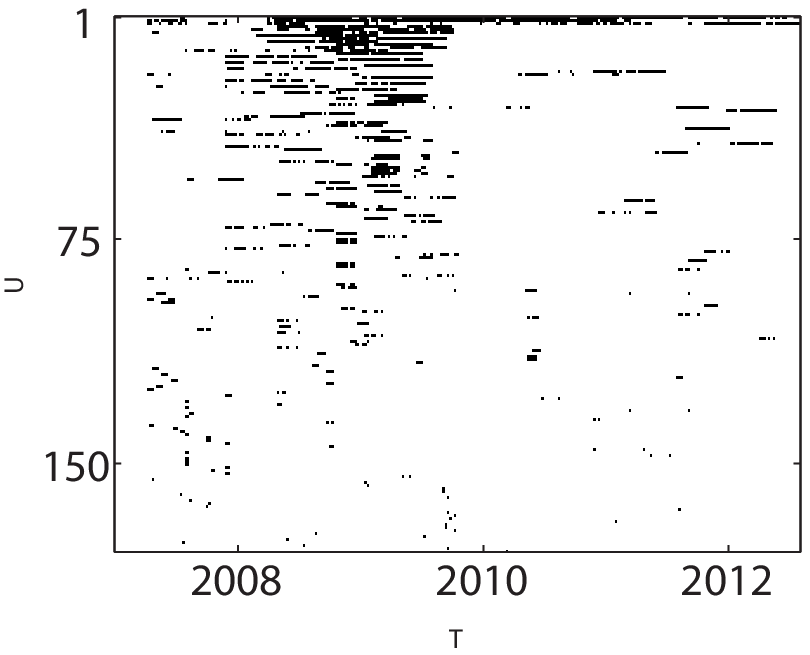}
\label{fig: geolifeact1}
}
\caption{(a) Gridding of the $5^{\text{th}}$ ring road of Beijing into squares of $100~\text{m} \times 100~\text{m}$. In area has approximate size of $39~\text{Km} \times 39~\text{Km}$. The grids in which a GPS position is recorded for a user is darkened. (b) The active weeks for each user during the data collection campaign.}
\end{figure}

The Geolife (GL) dataset~\cite{zheng2008understanding} contains the GPS traces of $182$ users collected over five years.
The user traces in this dataset are represented by a sequence of time-stamped points, each of which contains the information of latitude and longitude.
The trajectories are widely distributed over many cities in China and even some in  the USA and Europe, but the majority of the data is created in the city of Beijing.
%In the following we refer to this dataset by the GL dataset.
In our experiments, we focus on the trajectories collected within the $5^{\text{th}}$ ring road of Beijing, which is an area approximately $39~\text{km} \times 39~\text{km}$.
We first grid this area into $100~\text{m} \times 100~\text{m}$ squares.
Each square represents a location.
Figure~\ref{fig: geolifemap1} shows the considered area, where all the locations with a recorded GPS position are darkened.
We call a particular one-week period \emph{active} for a user if she has at least one recorded GPS position during the week.
Figure~\ref{fig: geolifeact1} shows the active weeks for each user during the data collection campaign.
As can be seen from Figure~\ref{fig: geolifeact1}, the users contributed to the dataset during different periods.

We filtered out all users with number of active weeks equal to $1$ and were left with  $\pop=154$ users.
The users have on average $15.4$ active weeks of data.
We split each user's trajectories into two parts, where part  one corresponds to the trajectories recorded in the first half of her active weeks, and part two corresponds to the trajectories  recorded in the second half of her active weeks.
%For example, if a user has trajectories recorded in weeks 10, 15, and 100; his trajectories will b%e split into part one (corresponding to trajectories recorded in weeks 10 and 15) and part two (corresponding to trajectories recorded in weeks 100).
We   construct  histograms of the locations visited by each user in each week.
Thus, the histogram $\Gamma_{x_{\permU(i)}}$ (respectively, $\Gamma_{y_i}$) of user $i$ in the first (respectively, second) part gives the relative fractions of recorded GPS positions from various locations (i.e., grid squares on the map) in the first (respectively, second) part of her data.
The set  $\SetSa$ (respectively, $\SetSb$) corresponds to the  histograms of the number of recorded GPS positions  of the $\pop$ users from the $\staten$ locations  in their first parts (respectively, second parts).
%We consider the scenario where call histograms in the form of $\SetSa$  have been released; and an attacker who has  access to auxiliary histograms in the form of $\SetSb$ tries to match the histograms that belong the same underlying user.
%Given $\SetSa$ and $\SetSb$, the goal of the matching task is to match every histogram in $\SetSb$ to the histogram in $\SetSa$ produced by the same user, or equivalently, to estimate the permutation $\permU$ in~\eqref{eq:Ch4perturb1}.
%\notes{first part and second part different for different users.}

\subsubsection{Matching accuracy with different metrics}\label{sec: geolifematchingexp0}
We set the side length of grid squares equal to $1000~\text{m}$, and we compute the histograms and apply (A1) to the graph $G$ of Figure~\ref{fig: bipartitematch4traj1} with different choices of edge weights.
%After computing the histograms, we construct the complete bipartite graph $G$ shown in Figure~\ref{fig: bipartitematch4traj1} for grid side-length equal to $1000~\text{m}$ . %as described in Section~\ref{sec: proposedapproach1}.
%We  apply (A1)  by using different choices of edge weights.
The obtained matching accuracy, when  the side length of grid squares is $1000~\text{m}$, is shown in the last row of Table~\ref{table:resDatsets}.
The accuracy obtained by using the weight function proposed in~\eqref{eqn:weights} is significantly higher than that obtained by using any of the other \emph{heuristic} measures.
% by using the proposed metric in~\eqref{eqn:weights}, $l_1$-norm in~\eqref{eq: Ch4cossim1}, cosine distance in~\eqref{eq: Ch4lpdist1}, and  the dot product in~\eqref{eq: Ch4dotsim1} was $90\%$, $83\%$, $77.5\%$, and $70\%$, respectively.
%We observe that   the matching accuracy obtained using weight function proposed in~\eqref{eqn:weights} is significantly higher than that obtained by using any of the other \emph{heuristic} measures.

%\notes{Should we use grid side-length or side-length of the grid squares?}

\subsubsection{Effect of spatial resolution}\label{sec: geolifematchingexp1}
\begin{figure}
\centering
\psfrag{p}[l][l]{\footnotesize{\text{proposed}}}
\psfrag{l}[l][l]{\footnotesize{\text{$l_1$-norm}}}
\psfrag{c}[l][l]{\footnotesize{\text{cosine}}}
\psfrag{d}[l][l]{\footnotesize{\text{dot}}}
\psfrag{A}[Bc][Bc]{\small{Accuracy ($\%$)}}
\psfrag{G}[Bc][Bc]{\small{Side lengths of grid squares (in meters)}}
\includegraphics[width=1\columnwidth]{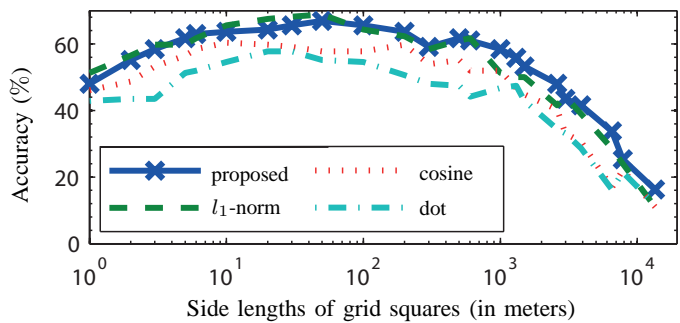}
\caption{The evolution of the matching accuracy for the GL dataset ($\pop=154$) as a function of the grid side-length by using different metrics.
%The measures are as follows: proposed~\eqref{eqn:weights}, $l_1$-norm~\eqref{eq: Ch4lpdist1}, cosine distance~\eqref{eq: Ch4cossim1}, and dot product~\eqref{eq: Ch4dotsim1}.
The measures are defined in~\eqref{eqn:weights},~\eqref{eq: Ch4lpdist1},~\eqref{eq: Ch4cossim1}, and~\eqref{eq: Ch4dotsim1}.
The accuracy is maximum for moderate side-lengths.
 %The proposed weight function yields the highest percentage of accuracy in the matching.
 }
\label{fig: geolifematchres1}
\end{figure}

%After computing the histograms, we construct the complete bipartite graph $G$ shown in Figure~\ref{fig: bipartitematch4traj1}.
%We apply (A1) to the graph by using different choices of edge weights.
We repeat the previous experiment with varying choices for the side lengths of grid squares.
The resulting matching accuracies are shown in Figure~\ref{fig: geolifematchres1} as a function of the side lengths.
For very large  side-lengths, the spatial resolution is low, hence  the users' location traces are easily confused, thus leading to low  matching accuracy.
%As the grid side-length increases (i.e., spatial resolution decreases) the quality of the data degrades and thus the matching accuracy decreases.
%There is  some inherent variability is the GPS trajectories because of  GPS errors and the minor variations in the user's locations
For very small side-lengths, there are too many locations in the sense that the inherent noise in the GPS trajectories come into effect, which leads to an over-fitting of the data, and thus the matching accuracy is again low.
 %some inherent variability due to GPS errors or also due to minor variations in the user's locations
 Therefore, the accuracy is maximum for moderate side-lengths -- around $100~\text{m}$ in the figure.
%We observe that   the matching accuracy obtained using weight function proposed in~\eqref{eqn:weights} is  higher than that obtained by using any of the other \emph{heuristic} measures when the grid side-length is greater than $100~\text{m}$.
%[RRRR: explanation of the decrease]

\subsubsection{Comparison with existing work}\label{sec: geolifecomprelated}
In~\cite{gambs2014anonymization} the authors propose a de-anonymization scheme  based on a mobility model called the Mobility Markov Chain (MMC) and applied it to the GL dataset.
%Here we discuss the matching accuracy obtained  by two existing de-anonymization methods in the literature on the geoLife dataset.
%The authors of~\cite{gambs2014anonymization} proposed a de-anonymization attack based on a mobility model called Mobility Markov Chain (MMC).
In their approach, an MMC is constructed for each user from her mobility traces observed during the training phase and  during the test phase.
Distance metrics  between MMCs are then used to link a user's trace from the test phase to the corresponding trace in the training phase.
There are three main differences between their approach  and ours.
First, in their approach, the set of locations that a user visits is learned by applying a clustering algorithm to the user's GPS trajectories.
The clustering algorithm identifies the accumulation regions of the user's trajectory that is then used to represent the set of locations that the user visits, whereas  
in our approach,  we partition the map area into squares that represent the set of locations.
Second, they use the timing information present in the users' trajectories to learn the MMCs, whereas in our case we disregard all the timing information present in the trajectories and only consider the fraction of visits to different locations.
Third, they de-anonymize the users one-by-one, whereas we simultaneously de-anonymize  all the users.

In~\cite{gambs2014anonymization}, the authors report a de-anonymization accuracy of up to $45\%$ on $77$ users in the setting where the regions identified from the clustering algorithm have a maximum radius equal to $500~\text{m}$.
In comparison, our scheme obtains a de-anonymization accuracy of up to $60\%$ for $154$ users in the setting where the side lengths of grid squares  range from $300~\text{m}$ to $1000~\text{m}$.
If we do one-by-one user de-anonymization, this accuracy drops down to $50\%$, however it still remains higher than the $45\%$ reported in~\cite{gambs2014anonymization}.
%[RRRR: explanation of why its higher!]
We believe that this is because by using a complicated and dynamic model such as MMC, there is a substantial over fitting of the user data to the model.
In~\cite{gambs2014anonymization}, a  $\staten \times \staten$  transition probability matrix is fitted to each trace, whereas in our approach a $\staten$-length probability vector is fitted.
%In other words, for each trace, a $\staten \times \staten$ matrix  is learned in~\cite{gambs2014anonymization}, whereas a $\staten$-length vector is learned in our approach.
This leads to  poorer performances because the model learned from the first dataset does not ``generalize'' well to the second dataset.

\subsection{Running time}\label{sec: expruntim1}
Here we present the timing information of performing the  de-anonymization attacks that are given in Table~\ref{table:resDatsets}. 
We consider only the case where our proposed metric is used.
The running times are given for MATLAB version $8.3.0.532$ running on a Lenovo Thinkpad T$410$ equipped with Intel i$7$ processor with clock speed of $2.67$ GHz, with $8$ Gb of RAM, and with Microsoft Windows $7$ $64$-bit operating system.

The running time for computing the edge weights ($w_{\z i}$ in~\eqref{eqn:weights}) of graph $G$  and for running (A1) on $G$ are $41~\text{min}$  and $432~\text{min}$, respectively,  for the CDR dataset.
The respective numbers for the WBH dataset are $6~\text{sec}$ and $0.1~\text{sec}$  for computing the edge weights of $G$  and for running (A1) on $G$, respectively, 
The respective numbers for the GL dataset are $0.9~\text{sec}$ and $0.2~\text{sec}$ for computing the edge weights of $G$  and for running (A1) on $G$, respectively.
Note that the reported numbers do not include the preprocessing time, that is, the time required for computing the histograms from the raw data.

%%%%%%%%%%%%%%%%%%%%%%%%%%%%%%%%%%%%%%%%%%%%%%%%%%%%%%%%%%%%%%%%%
%%%%%%%%%%%%%%%%%%%%%%%%%%%%%%%%%%%%%%%%%%%%%%%%%%%%%%%%%%%%%%%%%
%%%%%%%%%%%%%%%%%%%%%%%%%%%%%%%%%%%%%%%%%%%%%%%%%%%%%%%%%%%%%%%%%
%%%%%%%%%%%%%%%%%%%%%%%%%%%%%%%%%%%%%%%%%%%%%%%%%%%%%%%%%%%%%%%%%
\section{Privacy enhancing mechanisms}\label{sec: possiblesol}
%We demonstrated by our experiments in Section~\ref{sec: experiments} that anonymization of histograms of users' behavior is not sufficient to preserve the privacy of the users against an adversary who has access to auxiliary knowledge about the users.
We demonstrated by our experiments in Section~\ref{sec: experiments} that applying anonymization to histograms of users' behavior is not effective in protecting the users' identities from an adversary who has access to auxiliary knowledge about the users.
In this section, we discuss additional privacy-preserving mechanisms that can be applied to the histograms in order to make it difficult for the adversary to identify the users.
% reduce the de-anonymization accuracy of the adversary.
These mechanisms essentially make the released histograms closer to each other so that there is greater scope for confusion in  distinguishing them from each other, and thus the matching accuracy  declines. % it becomes more difficult to

\subsection{Basic data coarsening and data suppression}\label{sec: possiblesol0}
Two popular categories of privacy-preserving mechanisms  are \emph{data obfuscation} and \emph{data suppression} methods~\cite{aggarwal2008general}.
%Data obfuscation refers to the process of coarsening the  resolution  of the released histograms.
An example of data coarsening is spatial resolution reduction, which can be achieved by   aggregating different locations into one.
We investigated the latter in our experiments in Section~\ref{sec: experlocobfus1} and in Figure~\ref{fig: geolifematchres1}.
Data suppression is the process of restricting the released data associated with each user.
For example, in our experiment in Figure~\ref{fig: WBHpopularexp1} for the WBH dataset, we consider only the subset of  popular websites  (i.e., websites that are visited by \emph{most} users) and publish the histograms values associated with this subset.
Another example is time-domain restriction, which refers to the process of limiting the time-period over which the histograms are computed.
We investigated this approach in our experiment in Figure~\ref{fig: experdayvar1} for the CDR dataset.
Another popular privacy-preserving mechanism is $k$-anonymization, which we investigate in the next subsection.

\subsection{$k$-Anonymization via micro-aggregation}\label{sec: kanonysol}
%We demonstrated by our experiments in Section~\ref{sec: experiments} that applying anonymization to histograms of users' behavior is not effective against an adversary who has access to auxiliary knowledge about the users.
%In this subsection we discuss $k$-anonymization, which is a technique that can be applied to  the histograms in order to guaranty lower matching accuracy to the adversary.
%additional privacy preserving mechanisms that can be applied to the histograms in order to guaranty lower matching accuracy to the adversary.
%These mechanism essentially make the released histograms closer to each other so that it becomes more difficult to distinguish them, and thus the matching accuracy  declines.
A released dataset  is said to have the $k$-anonymity property if the data for each user contained in the  dataset is identical to the data for  at least $k-1$ other users~\cite{sweeney2002k}.
One mechanism for guaranteeing $k$-anonymity for a dataset  is by means of micro-aggregation~\cite{domingo2002practical}.
In micro-aggregation, users' data are partitioned into different clusters such that each cluster contains data of at least $k$ users.
The average of the  data within each cluster is computed and  then  used to replace the original data values of all the users within the cluster.
These new data values are then released, resulting in a dataset with the $k$-anonymity property.
In micro-aggregation, the partitioning is done by using a criterion of minimum within-cluster information loss, and it has been shown that finding the optimal partitioning  is NP-hard~\cite{domingo2008polynomial}.
In the following, we define micro-aggregation in mathematical terms, and describe how our matching method can be adapted to de-anonymize micro-aggregated  histograms of users' data.

\subsubsection{Micro-aggregation}\label{sec: microaggdef}
Let $\left\{C_1, C_2, \ldots, C_g\right\}$ be a partitioning of the users $\USetSa$ (i.e., the users who generate the histograms $\SetSa$) into $g$ clusters.
That is, $\USetSa=\cup_{q=1}^{g} C_q$ and $C_q \cap C_{q^\prime} =\emptyset$ for $q\neq q^\prime$.
We later elaborate on the criteria for choosing the set $\left\{C_q\right\}_{1 \leq q \leq g}$.
Furthermore, define $k=\min_{1 \leq q \leq g} |C_q|$, and
\begin{equation}
\Gamma_{C_q}=\frac{1}{|C_q|}\sum_{\z \in C_q }\Gamma_{x_\z},
\end{equation}
for $1 \leq q \leq g$, which represent the average of histograms of all users within each cluster.
In micro-aggregation, instead of releasing the set of histograms $\SetSa$, the set of micro-aggregated  histograms $\SetSaTilde=\left\{\widetilde{\Gamma}_{x_\z}\right\}$ is released, where  $\widetilde{\Gamma}_{x_\z}=\Gamma_{C_q}$ for $\z \in C_q$ and for $1\leq q \leq g$.
%Let $\tilde{\SetSa}=\left\{\tilde{\Gamma}_{x_z}\right\}$.
It is straightforward to see that  when  $\SetSaTilde$ is released, every user in set $\USetSa$ is guaranteed $k$-anonymity.

Although micro-aggregation  guarantees $k$-anonymity to the users, it distorts  the released dataset.
Specifically, every histogram  $\Gamma_{x_\z}$ is replaced by $\widetilde{\Gamma}_{x_\z}$.
The criteria for obtaining $\left\{C_q\right\}_{1 \leq q \leq g}$ in micro-aggregation is to minimize the total distortion to the data,  for a given value of $k$.
In the literature,  the $l_2$-norm is often used to measure the distortion~\cite{panagiotakis2013successive}, however because the histograms lie on the probability simplex, we use the $l_1$-norm to measure the distortion.
In particular the total added distortion, which is also called  \emph{information loss}, is $\sum_{q=1}^{g} \sum_{\z \in C_q} \norm{\Gamma_{x_\z} - \Gamma_{C_q}}_1.$
The maximum information loss occurs when all the users are partitioned into a single cluster, i.e., when $g=1$.
The information loss in this case is $\sum_{\z=1}^{\pop} \norm{\Gamma_{x_\z} - \overline{\Gamma}_{x}}_1$, where $\overline{\Gamma}_{x}=\sum_{\z=1}^{\pop}\Gamma_{x_\z}/\pop$.
Consequently, we can define a normalized information loss measure as follows:
\begin{equation}\label{eq: norinfoloss1}
L=\sum_{q=1}^{g} \sum_{\z \in C_q} \norm{\Gamma_{x_\z} - \Gamma_{C_q}}_1 \left/ \sum_{\z=1}^{\pop} \norm{\Gamma_{x_\z} - \overline{\Gamma}_{x}}_1 \right..
\end{equation}
The extreme case, $L=0$, represents the scenario where no micro-aggregation is performed (i.e., $g=\pop$) and where all users are guaranteed $1$-anonymity.
The other extreme case,  $L=1$, represents the scenario where $g=1$ and where all users are guaranteed $\pop$-anonymity.
For a given value of $k$, we seek the partitioning $\left\{C_q\right\}_{1 \leq q \leq g}$ whose normalized information loss $L$ given in~\eqref{eq: norinfoloss1} is as small as possible.
%In our experiment in the following, we use the algorithm proposed in~\cite{panagiotakis2013successive} which is a polynomial approximation to the micro-aggregation problem.
In our following experiment, we use the algorithm proposed in~\cite{panagiotakis2013successive} for performing micro-aggregation, where we adapt the algorithm to measure the distortion by using $l_1$-norm.

\subsubsection{Experimental evaluations}\label{sec: kanonyexpeval}
Here we evaluate the effectiveness of the matching algorithm when  micro-aggregation is performed on the unlabeled histograms $\SetSa$. % in order to further protect the privacy of  the users by guaranteeing them $k$-anonymity.
%The released histograms $\SetSaTilde$ are thus  grouped into different clusters such that each cluster has  minimum size $k$ and that all the histograms in each cluster have identical values.\notes{Description not consistent with previous paragraph}
We consider an adversary who has access to the labeled histograms  $\SetSb$ and is interested in matching these histograms to the micro-aggregated ones in  $\SetSaTilde$.
We consider two different notions of accuracy  for the matching.
Let the labeled histogram $\Gamma_{y_i}$ be matched to the unlabeled micro-aggregated histogram $\widetilde{\Gamma}_{x_\z}$.
According to our first notion, there is a correct match if  $\z=\permU(i)$, where $\permU$ is defined in~\eqref{eq:Ch4perturb1}.
According to our second notion, there is a correct match if  $\widetilde{\Gamma}_{x_\z}=\widetilde{\Gamma}_{x_{\permU(i)}}$.
The former notion  of accuracy (called \emph{user-level}) measures the number of correctly matched users and is the same notion that  we used in  our experiments in Section~\ref{sec: experiments}, whereas the latter notion (called \emph{cluster-level})  measures the number of users whose $k$-anonymity class (i.e., cluster) is correctly identified.

For the CDR dataset, we consider the setting described in Section~\ref{sec: expervaryingN1}.
In particular,  we randomly choose $\pop=1000$ out of the 46986 users and construct the sets   $\SetSa$ and  $\SetSb$.
%, which contain the histograms of the proportion of calls made by the $\pop$ users  from different GSM antennas during the first week and the second week, respectively.
For the WBH dataset, we consider  the subset of the top $\staten=100$ popular websites and construct the sets of histograms  $\SetSa$ and  $\SetSb$ as described in Section~\ref{sec: WBHpopularexp1}.
%The sets   $\SetSa$ and  $\SetSb$ contain the histograms of the visits of the $\pop=102$ users to these websites during the first week and the second week, respectively.
For the GL dataset, we consider the setting described in Section~\ref{sec: geolifematchingexp0}  when grid side-length is set equal to $1000~\text{m}$.
%The sets   $\SetSa$ and  $\SetSb$ contain the histograms of the recorded GPS positions  of the $\pop=154$ users in different locations during the first part and the second part, respectively.

\begin{figure}
\centering
\subfigure[CDR dataset ($\pop=1000$)]{
\psfrag{u}[l][Bc]{\scriptsize{U-Lev. accuracy}}
\psfrag{c}[l][Bc]{\scriptsize{C-Lev. accuracy}}
\psfrag{i}[l][Bc]{\scriptsize{Information loss}}
\psfrag{n}[l][Bc]{\scriptsize{$\#$ clusters}}
\psfrag{k}[Bc][Bc]{\small{$k$}}
\psfrag{p}[Bc][Bc]{\small{$\%$}}
\includegraphics[width=.6\columnwidth]{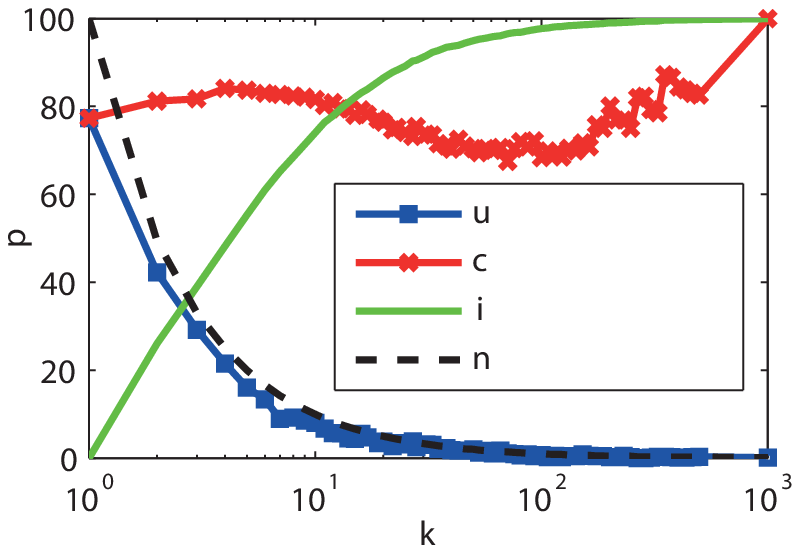}
\label{fig: microAggCDR}
}
\hfill
\subfigure[WBH dataset ($\pop=102$)]{
\psfrag{k}[Bc][Bc]{\small{$k$}}
\psfrag{p}[Bc][Bc]{\small{$\%$}}
\includegraphics[width=.46\columnwidth]{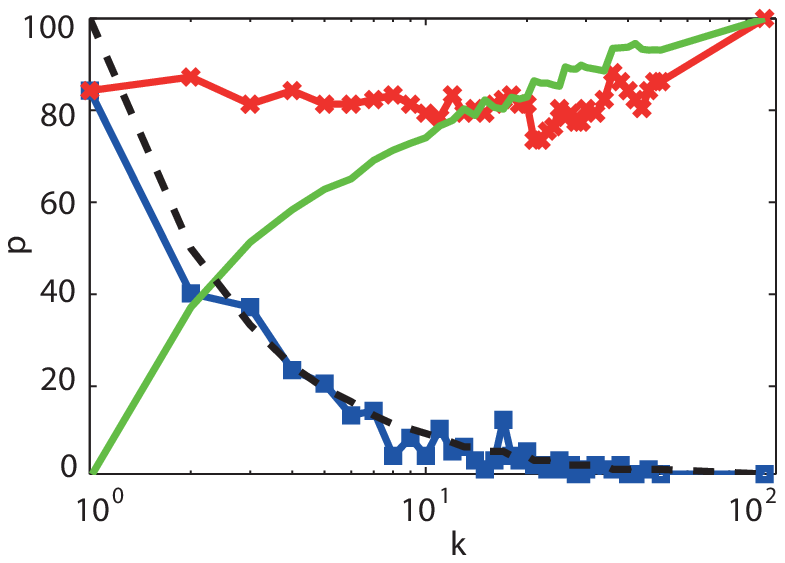}
\label{fig: microAggWBH}
}
\subfigure[GL dataset ($\pop=154$)]{
\psfrag{k}[Bc][Bc]{\small{$k$}}
\psfrag{p}[Bc][Bc]{\small{$\%$}}
\includegraphics[width=.46\columnwidth]{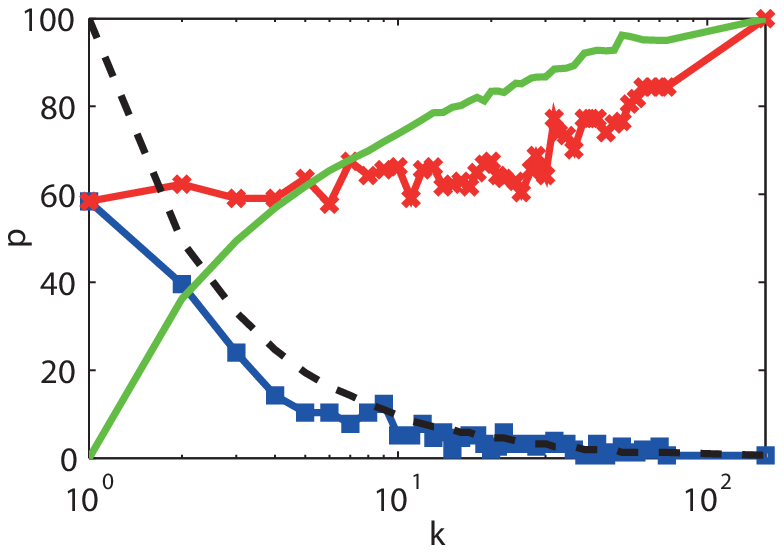}
\label{fig: microAgggeo}
}
\caption{The trade-off between user-level (denoted by U-Lev.) and cluster-level (denoted  by C-Lev.) matching accuracies and the information loss $L$ as $k$-anonymity is guaranteed to the users. As $k$ increases, more distortion is added to the histograms (i.e., more information is lost) but the user-level accuracy drops meaning that the users enjoy higher privacy with respect to the adversary. The cluster-level accuracy however experiences much less fluctuation.}
\label{fig: microAgg}
\vspace{-5pt}
\end{figure}

For each dataset, we perform micro-aggregation with different values of $k$ on the set $\SetSa$.
We then perform the matching between $\SetSaTilde$ and $\SetSb$ by using only the proposed metric of~\eqref{eqn:weights}.
The obtained accuracies are shown in Figure~\ref{fig: microAggCDR} and~\ref{fig: microAggWBH} and~\ref{fig: microAgggeo} for the CDR, WBH, and  GL dataset, respectively.
The figures also show the normalized information loss $L$ defined in~\eqref{eq: norinfoloss1} and the normalized number of clusters (i.e, $g/N$), expressed in percentages.
In the extreme case with  $k=1$, no micro-aggregation is performed; therefore, $g=\pop$, $L=0$, and the user-level accuracy is equal to the cluster-level  accuracy.
In the other extreme case,  $k=\pop$, and all the released unlabeled histograms are identical;  therefore, the information loss is maximum ($L=1$), and while the user-level accuracy is minimum, the cluster-level accuracy is maximum.
As $k$ increases to about $10$, the user-level accuracy dramatically drops, hence the users enjoy an increased level of privacy  guarantee, whereas the cluster-level accuracy remains almost the same for all values of $k$.
%[RRRR: story improvement needed!]

%%%%%%%%%%%%%%%%%%%%%%%%%%%%%%%%%%%%%%%%%%%%%%%%%%%%%%%%%%%%%%%%%
%%%%%%%%%%%%%%%%%%%%%%%%%%%%%%%%%%%%%%%%%%%%%%%%%%%%%%%%%%%%%%%%%
%%%%%%%%%%%%%%%%%%%%%%%%%%%%%%%%%%%%%%%%%%%%%%%%%%%%%%%%%%%%%%%%%
%%%%%%%%%%%%%%%%%%%%%%%%%%%%%%%%%%%%%%%%%%%%%%%%%%%%%%%%%%%%%%%%%

\section{Conclusion}\label{sec: conc}
We have studied the task of identifying users from the statistics of their behavioral patterns.
%matching data strings that are generated by the same user from two different datasets.
%user identities across two datasets when the information available about the users is in the form of histograms.
%In this paper  we have studied strategies for de-anonymizing anonymized user histograms given auxiliary information about the user's behavior.
%We obtained an asymptotically optimal strategy for this problem assuming an i.i.d. model for the users' data.
Specifically, given an anonymized dataset in the form of histograms belonging to a set of users  and another independent set of histograms generated by the same set of users, we have shown that it is possible to identify the identities of the users in the first dataset to a surprising level of accuracy by matching the statistical characteristics of the users' behaviors across the two datasets.
Thus data histograms act as \emph{fingerprints} for identifying users.
%Hence,  statistics about users' behaviors  can act as users' fingerprints.
%We showed that two sets of histograms belonging
%
%We focused on the setting in which we are given two sets of labeled and unlabeled histograms and considered the task of an adversary seeks to  match the histograms that are generated by the same user.
Our proposed solution can be implemented via a minimum-weight maximal matching algorithm on a  complete  weighted bipartite graph and yields higher accuracy than heuristics-based methods on three different datasets of different nature.
%We studied the effect of various
We have studied the performance of the algorithm over a wide range of experimental conditions and demonstrated the effect of various factors, such as the number of users, the resolution of the data, the duration of the data collection, and the amount of data suppressed, on the accuracy of the matching algorithm.
We have gained the insight that  the simultaneous matching of the users yields higher accuracy compared to one-by-one user matching.
Furthermore, we have demonstrated the power of simplicity of statistics: Identification based only on data statistics can sometimes result in higher accuracy than existing methods based on more complicated data models.
We have further studied the performance of the algorithm under privacy-enhancement techniques, such as $k$-anonymization, and demonstrated the effect of $k$ on the matching accuracy.
Our results suggest that users can be identified, to a surprisingly high level of accuracy, even from the statistics of their behavior.
Moreover,  using the correct metric and optimal matching algorithm can lead to a significant improvement in matching accuracy over heuristics-based methods.
Privacy enhancement via $k$-anonymization and data obfuscation can reduce identification accuracy, but the accuracy remains non-negligible for moderate levels of data distortion.

\section{Acknowledgments}\label{sec: ack}
We thank Jacques Raguenez and Vincent Blondel for granting us access to the Orange CDR dataset. We thank Alessandra Sala for her  feedback on the manuscript.

%%%%%%%%%%%%%%%%%%%%%%%%%%%%%%%%%%%%%%%%%%%%%%%%%%%%%%%%%%%%%%%%%%%%%%%%%%%%%%%%%%%
%%%%%%%%%%%%%%%%%%%%%%%%%%%%%%%%%%%%%%%%%%%%%%%%%%%%%%%%%%%%%%%%%%%%%%%%%%%%%%%%%%%
%%%%%%%%%%%%%%%%%%%%%%%%%%%%%%%%%%%%%%%%%%%%%%%%%%%%%%%%%%%%%%%%%%%%%%%%%%%%%%%%%%%
%%%%%%%%%%%%%%%%%%%%%%%%%%%%%%%%%%%%%%%%%%%%%%%%%%%%%%%%%%%%%%%%%%%%%%%%%%%%%%%%%%%
%%%%%%%%%%%%%%%%%%%%%%%%%%%%%%%%%%%%%%%%%%%%%%%%%%%%%%%%%%%%%%%%%%%%%%%%%%%%%%%%%%%

%\section*{Acknowledgment}

\ifCLASSOPTIONcaptionsoff
  \newpage
\fi



\bibliographystyle{IEEEtran}
\bibliography{privrefs}

\begin{thebibliography}{10}
\providecommand{\url}[1]{#1}
\csname url@samestyle\endcsname
\providecommand{\newblock}{\relax}
\providecommand{\bibinfo}[2]{#2}
\providecommand{\BIBentrySTDinterwordspacing}{\spaceskip=0pt\relax}
\providecommand{\BIBentryALTinterwordstretchfactor}{4}
\providecommand{\BIBentryALTinterwordspacing}{\spaceskip=\fontdimen2\font plus
\BIBentryALTinterwordstretchfactor\fontdimen3\font minus
  \fontdimen4\font\relax}
\providecommand{\BIBforeignlanguage}[2]{{%
\expandafter\ifx\csname l@#1\endcsname\relax
\typeout{** WARNING: IEEEtran.bst: No hyphenation pattern has been}%
\typeout{** loaded for the language `#1'. Using the pattern for}%
\typeout{** the default language instead.}%
\else
\language=\csname l@#1\endcsname
\fi
#2}}
\providecommand{\BIBdecl}{\relax}
\BIBdecl

\bibitem{unniMNainiAlert13}
J.~Unnikrishnan and F.~Movahedi~Naini, ``De-anonymizing private data by
  matching statistics,'' in \emph{51th Annual Allerton Conference}, 2013.

\bibitem{unn13}
J.~Unnikrishnan, ``Asymptotically optimal matching of multiple sequences to
  source distributions and training sequences,'' \emph{IEEE Trans. Inf.
  Theory}, vol.~61, 2015.

\bibitem{resvar97}
P.~Resnick and H.~R. Varian, ``Recommender systems,'' \emph{Communications of
  the ACM}, vol.~40, no.~3, pp. 56--58, 1997.

\bibitem{Narayanan:2008:RDL:1397759.1398064}
A.~Narayanan and V.~Shmatikov, ``Robust de-anonymization of large sparse
  datasets,'' in \emph{Proceedings of IEEE Symposium on Security and Privacy},
  2008.

\bibitem{elmagarmid2007duplicate}
A.~K. Elmagarmid, P.~G. Ipeirotis, and V.~S. Verykios, ``Duplicate record
  detection: A survey,'' \emph{IEEE Trans. Knowl. Data Eng.}, 2007.

\bibitem{kalashnikov2008web}
D.~V. Kalashnikov, Z.~Chen, S.~Mehrotra, and R.~Nuray-Turan, ``Web people
  search via connection analysis,'' \emph{IEEE Trans. Knowl. Data Eng.}, 2008.

\bibitem{bengtson2008understanding}
E.~Bengtson and D.~Roth, ``Understanding the value of features for coreference
  resolution,'' in \emph{EMNLP Conference}, 2008.

\bibitem{stolerman2014breaking}
A.~Stolerman, R.~Overdorf, S.~Afroz, and R.~Greenstadt, ``Breaking the
  closed-world assumption in stylometric authorship attribution,'' in
  \emph{Advances in Digital Forensics X}.\hskip 1em plus 0.5em minus
  0.4em\relax Springer, 2014.

\bibitem{afroz2014doppelganger}
S.~Afroz, A.~C. Islam, A.~Stolerman, R.~Greenstadt, and D.~McCoy,
  ``Doppelg{\"a}nger finder: Taking stylometry to the underground,'' in
  \emph{Security and Privacy (SP), 2014 IEEE Symposium on}.\hskip 1em plus
  0.5em minus 0.4em\relax IEEE, 2014.

\bibitem{peled2013entity}
O.~Peled, M.~Fire, L.~Rokach, and Y.~Elovici, ``Entity matching in online
  social networks,'' in \emph{IEEE SocialCom}, 2013.

\bibitem{liu2013s}
J.~Liu, F.~Zhang, X.~Song, Y.-I. Song, C.-Y. Lin, and H.-W. Hon, ``What's in a
  name?: an unsupervised approach to link users across communities,'' in
  \emph{ACM WSDM}, 2013.

\bibitem{sweeney1997weaving}
L.~Sweeney, ``Weaving technology and policy together to maintain
  confidentiality,'' \emph{The Journal of Law, Medicine \& Ethics}, 1997.

\bibitem{zang2011anonymization}
H.~Zang and J.~Bolot, ``Anonymization of location data does not work: A
  large-scale measurement study,'' in \emph{ACM MobiCom}, 2011.

\bibitem{xiao2010finding}
X.~Xiao, Y.~Zheng, Q.~Luo, and X.~Xie, ``Finding similar users using
  category-based location history,'' in \emph{ACM SIGSPATIAL}, 2010.

\bibitem{ma2010privacy}
C.~Y. Ma, D.~K. Yau, N.~K. Yip, and N.~S. Rao, ``Privacy vulnerability of
  published anonymous mobility traces,'' in \emph{MobiCom}, 2010.

\bibitem{freudiger2012evaluating}
J.~Freudiger, R.~Shokri, and J.-P. Hubaux, ``Evaluating the privacy risk of
  location-based services,'' in \emph{Financial Cryptography and Data
  Security}.\hskip 1em plus 0.5em minus 0.4em\relax Springer, 2012.

\bibitem{gambs2014anonymization}
S.~Gambs, M.-O. Killijian, and M.~N{\'u}{\~n}ez~del Prado~Cortez,
  ``De-anonymization attack on geolocated data,'' \emph{Journal of Computer and
  System Sciences}, 2014.

\bibitem{de2008identification}
Y.~De~Mulder, G.~Danezis, L.~Batina, and B.~Preneel, ``Identification via
  location-profiling in gsm networks,'' in \emph{ACM WPES}, 2008.

\bibitem{machanavajjhala2008privacy}
A.~Machanavajjhala, D.~Kifer, J.~Abowd, J.~Gehrke, and L.~Vilhuber, ``Privacy:
  Theory meets practice on the map,'' in \emph{IEEE ICDE}, 2008.

\bibitem{olejnik2014uniqueness}
L.~Olejnik, C.~Castelluccia, and A.~Janc, ``On the uniqueness of web browsing
  history patterns,'' \emph{annals of telecommunications-annales des
  t{\'e}l{\'e}communications}, vol.~69, no. 1-2, pp. 63--74, 2014.

\bibitem{yen2012host}
T.-F. Yen, Y.~Xie, F.~Yu, R.~P. Yu, and M.~Abadi, ``Host fingerprinting and
  tracking on the web: Privacy and security implications.'' in \emph{NDSS},
  2012.

\bibitem{sharad2013anonymizing}
K.~Sharad and G.~Danezis, ``De-anonymizing d4d datasets,'' in \emph{HotPETs},
  2013.

\bibitem{srivatsa2012deanonymizing}
M.~Srivatsa and M.~Hicks, ``Deanonymizing mobility traces: Using social network
  as a side-channel,'' in \emph{ACM CCS}, 2012.

\bibitem{monhidverblo13}
Y.-A. de~Montjoye, C.~A. Hidalgo, M.~Verleysen, and V.~D. Blondel, ``{Unique in
  the Crowd: The privacy bounds of human mobility},'' \emph{Scientific
  Reports}, 2013.

\bibitem{shokri2011quantifying}
R.~Shokri, G.~Theodorakopoulos, J.-Y. Le~Boudec, and J.-P. Hubaux,
  ``Quantifying location privacy,'' in \emph{Security and Privacy (SP), 2011
  IEEE Symposium on}.\hskip 1em plus 0.5em minus 0.4em\relax IEEE, 2011, pp.
  247--262.

\bibitem{danezis2009vida}
G.~Danezis and C.~Troncoso, ``Vida: How to use bayesian inference to
  de-anonymize persistent communications,'' in \emph{Privacy Enhancing
  Technologies}.\hskip 1em plus 0.5em minus 0.4em\relax Springer, 2009, pp.
  56--72.

\bibitem{troncoso2008perfect}
C.~Troncoso, B.~Gierlichs, B.~Preneel, and I.~Verbauwhede, ``Perfect matching
  disclosure attacks,'' in \emph{Privacy Enhancing Technologies}.\hskip 1em
  plus 0.5em minus 0.4em\relax Springer, 2008, pp. 2--23.

\bibitem{pedarsani2011privacy}
P.~Pedarsani and M.~Grossglauser, ``On the privacy of anonymized networks,'' in
  \emph{ACM KDD}, 2011.

\bibitem{hastibfri09}
T.~Hastie, R.~Tibshirani, and J.~Friedman, \emph{The elements of statistical
  learning}.\hskip 1em plus 0.5em minus 0.4em\relax Springer, 2009, vol.~2,
  no.~1.

\bibitem{covtho06}
T.~M. Cover and J.~A. Thomas, \emph{Elements of Information Theory 2nd
  Edition}.\hskip 1em plus 0.5em minus 0.4em\relax Wiley-Interscience, 2006.

\bibitem{ramshaw2012weight}
L.~Ramshaw and R.~E. Tarjan, ``A weight-scaling algorithm for min-cost
  imperfect matchings in bipartite graphs,'' in \emph{IEEE FOCS}, 2012.

\bibitem{fredman1987fibonacci}
M.~L. Fredman and R.~E. Tarjan, ``Fibonacci heaps and their uses in improved
  network optimization algorithms,'' \emph{JACM}, 1987.

\bibitem{coocunpulsch11}
W.~Cook, W.~Cunningham, W.~Pulleyblank, and A.~Schrijver, \emph{Combinatorial
  Optimization}, ser. Wiley Series in Discrete Mathematics and
  Optimization.\hskip 1em plus 0.5em minus 0.4em\relax Wiley, 2011.

\bibitem{Duan:2014:LAM:2578041.2529989}
R.~Duan and S.~Pettie, ``Linear-time approximation for maximum weight
  matching,'' \emph{J. ACM}, vol.~61, no.~1, pp. 1:1--1:23, Jan. 2014.

\bibitem{blondel2012data}
V.~D. Blondel, M.~Esch, C.~Chan, F.~Clerot, P.~Deville, E.~Huens, F.~Morlot,
  Z.~Smoreda, and C.~Ziemlicki, ``Data for development: the d4d challenge on
  mobile phone data,'' \emph{arXiv preprint arXiv:1210.0137}, 2012.

\bibitem{huang2010preserving}
K.~L. Huang, S.~S. Kanhere, and W.~Hu, ``Preserving privacy in participatory
  sensing systems,'' \emph{Computer Communications}, 2010.

\bibitem{christin2011survey}
D.~Christin, A.~Reinhardt, S.~S. Kanhere, and M.~Hollick, ``A survey on privacy
  in mobile participatory sensing applications,'' \emph{Journal of Systems and
  Software}, 2011.

\bibitem{herder2011experiences}
E.~Herder, R.~Kawase, and G.~Papadakis, ``Experiences in building the public
  web history repository,'' in \emph{Proc. of Datatel Workshop}, 2011.

\bibitem{zheng2008understanding}
Y.~Zheng, Q.~Li, Y.~Chen, X.~Xie, and W.-Y. Ma, ``Understanding mobility based
  on gps data,'' in \emph{ACM Ubicomp}, 2008.

\bibitem{aggarwal2008general}
C.~C. Aggarwal and S.~Y. Philip, \emph{A general survey of privacy-preserving
  data mining models and algorithms}.\hskip 1em plus 0.5em minus 0.4em\relax
  Springer, 2008.

\bibitem{sweeney2002k}
L.~Sweeney, ``k-anonymity: A model for protecting privacy,'' \emph{INT J
  UNCERTAIN FUZZ}, 2002.

\bibitem{domingo2002practical}
J.~Domingo-Ferrer and J.~M. Mateo-Sanz, ``Practical data-oriented
  microaggregation for statistical disclosure control,'' \emph{IEEE Trans.
  Knowl. Data Eng.}, 2002.

\bibitem{domingo2008polynomial}
J.~Domingo-Ferrer, F.~Seb{\'e}, and A.~Solanas, ``A polynomial-time
  approximation to optimal multivariate microaggregation,'' \emph{Computers \&
  Mathematics with Applications}, 2008.

\bibitem{panagiotakis2013successive}
C.~Panagiotakis and G.~Tziritas, ``Successive group selection for
  microaggregation,'' \emph{IEEE Trans. Knowl. Data Eng.}, 2013.

\end{thebibliography}

\vfill


\end{document}